\documentclass[10pt,twocolumn,letterpaper]{article}

\usepackage[pagenumbers]{cvpr} %

\usepackage{url}
\usepackage{booktabs}
\usepackage{multirow}
\usepackage{multicol}
\usepackage{kotex}

\usepackage{algorithm}
\usepackage{listings}
\usepackage{xspace}
\usepackage{graphicx}
\usepackage{wrapfig}
\usepackage{subcaption}
\usepackage{pifont}

\definecolor{lightgreen}{RGB}{220,255,180}
\definecolor{darkergreen}{RGB}{21, 152, 56}
\newcommand\greenp[1]{\textcolor{darkergreen}{(#1)}}
\definecolor{red2}{RGB}{252, 54, 65}
\newcommand\redp[1]{\textcolor{red2}{(#1)}}
\newcommand\greenpscript[1]{\scriptsize\greenp{#1}}
\newcommand\redpscript[1]{\scriptsize\redp{#1}}

\newcommand{\ours}{MaskSub\xspace}

\usepackage{graphicx}
\usepackage{booktabs}
\usepackage{colortbl}
\usepackage{bm}

\definecolor{cvprblue}{rgb}{0.21,0.49,0.74}
\usepackage[pagebackref,breaklinks,colorlinks,allcolors=cvprblue]{hyperref}

\def\paperID{209} %
\def\confName{CVPR}
\def\confYear{2025}

\title{Masking meets Supervision: A Strong Learning Alliance}

\author{
Byeongho Heo \quad 
Taekyung Kim \quad 
Sangdoo Yun \quad 
Dongyoon Han \\ [1mm]
NAVER AI Lab
}

\begin{document}
\maketitle

\begin{abstract}
Pre-training with random masked inputs has emerged as a novel trend in self-supervised training. However, supervised learning still faces a challenge in adopting masking augmentations, primarily due to unstable training. In this paper, we propose a novel way to involve masking augmentations dubbed Masked Sub-branch (MaskSub). MaskSub consists of the main-branch and sub-branch, the latter being a part of the former. The main-branch undergoes conventional training recipes, while the sub-branch merits intensive masking augmentations, during training. MaskSub tackles the challenge by mitigating adverse effects through a relaxed loss function similar to a self-distillation loss. Our analysis shows that MaskSub improves performance, with the training loss converging faster than in standard training, which suggests our method stabilizes the training process. We further validate MaskSub across diverse training scenarios and models, including DeiT-III training, MAE finetuning, CLIP finetuning, BERT training, and hierarchical architectures (ResNet and Swin Transformer). Our results show that MaskSub consistently achieves impressive performance gains across all the cases. MaskSub provides a practical and effective solution for introducing additional regularization under various training recipes. Code available at \url{https://github.com/naver-ai/augsub}
\end{abstract}

\section{Introduction}

Supervised learning is the most basic and effective way to train a network to achieve high performance on a target task.
To improve supervised learning, diverse regularizations are developed and used as training recipes~\cite{touvron2021deit,touvron2022deit3,wightman2021resnet}, which represent a group of sophisticatedly tuned regularizations to maximize learning performance.
Supervised learning has always held an advantage over self-supervised learning~\cite{mocov3,caron2021dino} based on the benefit of supervision.
However, emergence of Vision Transformer (ViT)~\cite{dosovitskiy2020vit} and Masked Image Modeling (MIM)~\cite{bao2021beit,he2022masked,peng2022beit2} is changing this trends.
ViT, which lacks inductive bias compared to convolution networks, poses many challenges to generalization performance for supervised learning.
On the other hand, MIMs such as MAE~\cite{he2022masked} rise as an alternative pretraining method for ViT by achieving competitive performance with supervised learning recipes.
Although a recent study~\cite{touvron2022deit3} shows that new supervised learning outperforms MIMs, the gap is insignificant. Thus, MIMs are still a strong competitor of supervised learning methods.

MIM masks random areas of an input image and forces the network to infer the masked area using the remaining area.
A representative part of MIM is high mask ratios over 50\%.
Although MIM also works at small mask ratios, it shows remarkable performance when trained with a high mask ratio.
The high mask ratio is a major difference between MIM and supervised learning since this high mask ratio, over 50\%, is not beneficial in supervised learning.
Supervised learning also has utilized random masking as an augmentation~\cite{ghiasi2018dropblock,zhong2020randomerase}, but it significantly degrades performance when the masking ratio is high.
In other words, supervised learning is not applicable for strong masking augmentation.
We conjecture that it is a major problem of the current supervised learning recipe, and there is room for improvement by enabling strong masking.
\begin{figure*}[t]
\small
\centering
\includegraphics[width=0.9\textwidth]{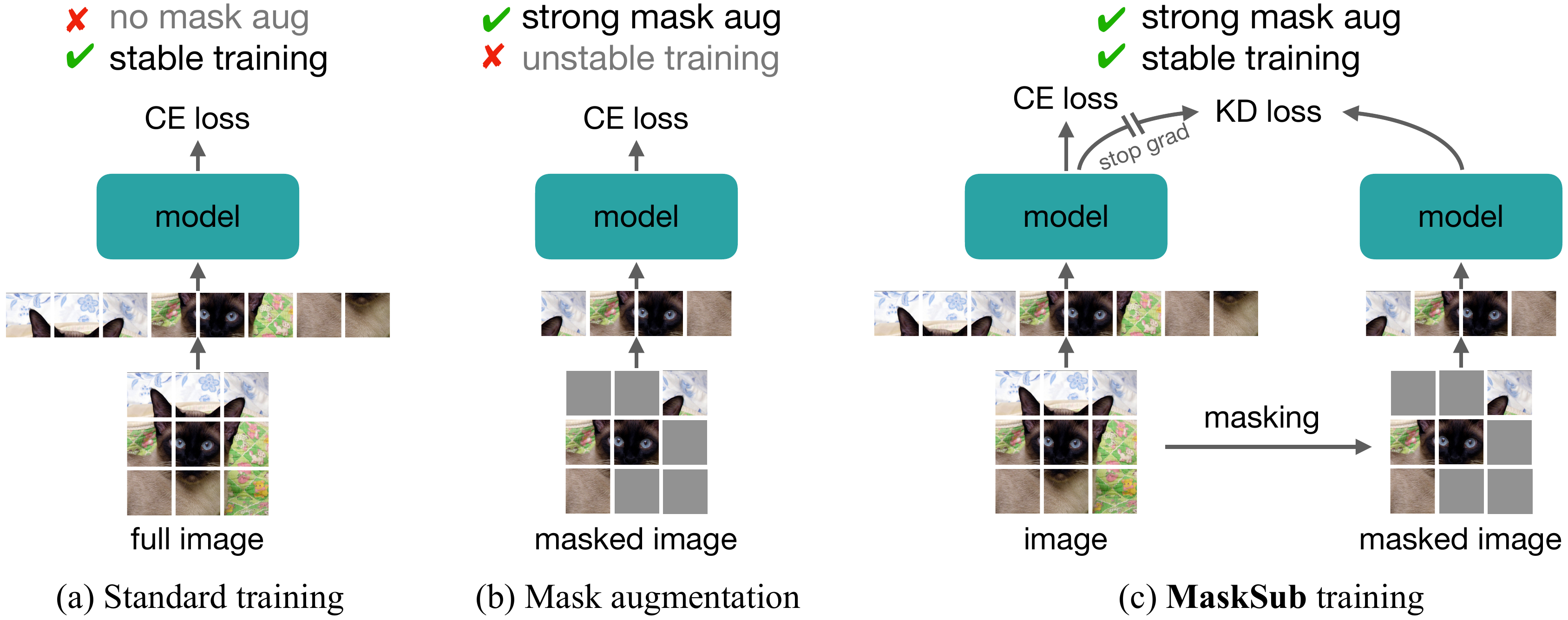}
\vspace{-.5em}
\caption{\textbf{Overview of Masked Sub-branch (MaskSub).} (a) standard supervised training; (b) masking augmentation training. The masking is applied to the main model, which degrades performance; (c) our \textbf{MaskSub} training, which separates the masking from the main model using the sub-branch and relaxes loss with self-distillation. MaskSub substantially improves the state-of-the-art training recipes~\cite{touvron2022deit3,wightman2021resnet}.}
\label{fig:title}
\vspace{-1em}
\end{figure*}

Our goal is to improve supervised learning with strong mask augmentation over 50\%. %
To this end, we introduce a novel learning framework using a ``sub-branch'' alongside the main-branch; throughout this paper, we use the term ``sub-branch'' to describe a model with dropped inputs.
The main-branch uses standard training recipes~\cite{wightman2021resnet,touvron2022deit3}, while the sub-branch utilizes mask augmentation. 
We name our method as Masked Sub-branch (MaskSub). 

We visualize the overview of MaskSub in Figure~\ref{fig:title}. 
We consider a high masking ratio over 50\% as similar in MAE~\cite{he2022masked}. 
Figure~\ref{fig:title} (b) shows that applying the strong random masking on the main-branch may lead to degraded performance.  
In contrast, as in Figure~\ref{fig:title} (c), MaskSub leverages the sub-branch for random masking, and the sub-branch receives the training signal from the main-branch similar to the self-distillation~\cite{one,selfdistill1,multi_exit}.
While the random masking technique amplifies the difficulty of the training process, this is counterbalanced by self-distillation loss since the outputs of the main-branch are relaxed and easy-to-learn objective than the ground-truth label. In summary, MaskSub applies a mask augmentation separated from the main-branch, utilizing a relaxed loss form.

We analyze MaskSub using 100 epochs training on ImageNet~\cite{deng2009imagenet}. 
Without MaskSub, loss convergence speed and corresponding accuracy are significantly degraded when mask augmentation is applied.
Conversely, MaskSub mitigates potential harmful effects from additional regularization, leading to a network training process that is even more efficient than standard training procedures.
Also, MaskSub is not limited to mask augmentation and can be used for general drop regularizations.
As a result, MaskSub is expanded to any random drop regularization without disrupting the convergence of original train loss; we employ three in-network drop-based options to show the applicability: masking~\cite{he2022masked,bao2021beit}, dropout~\cite{srivastava2014dropout}, and drop-path~\cite{huang2016stodepth,fan2019structuredrop}. 
Corresponding to each respective regularization strategy, we denote them MaskSub, DropSub, and PathSub.
Among the three variants, MaskSub notably exhibits a remarkable performance enhancement, demonstrating the necessity of mask augmentation in supervised learning.

We extensively validate the performance of MaskSub. MaskSub is applied on various state-of-the-art supervised learnings including DeiT-III training~\cite{touvron2022deit3}, MAE finetuning~\cite{he2022masked}, BEiTv2 finetuning~\cite{peng2022beit2}, CLIP finetuning~\cite{dong2022ftclip}, BERT training~\cite{devlin2018bert}, ResNet-RSB~\cite{wightman2021resnet}, and Swin transformer~\cite{liu2021swin}.
MaskSub demonstrates remarkable performance improvement in all benchmarks. 
We argue that MaskSub can be regarded as a novel way to utilize regularization for visual recognition.

\section{Related Work}
\textbf{Training recipe} has been considered an important ingredient in building a high-performance network. 
He et al.~\cite{he2019bag} demonstrate that the training recipe significantly influences the network performance.
RSB~\cite{wightman2021resnet} is a representative and high-performance recipe for ResNet.
With the emergence of ViT~\cite{dosovitskiy2020vit}, the training recipe for ViT has gained the attention of the field.
DeiT~\cite{touvron2021deit} shows that ViT can be trained to display strong performance with only ImageNet-1k~\cite{deng2009imagenet}.
DeiT-III~\cite{touvron2022deit3} is an improved version of DeiT, which applies findings from RSB to DeiT instead of distillation from CNN teacher. 
It is challenging to implement stronger or additional regularization in existing training recipes. To address this issue, we propose our MaskSub employing sub-branchs.

CoSub~\cite{touvron2022co} introduces a similar concept to ours, utilizing sub-branchs.
However, the sub-branch objective differs: while MaskSub aims to stabilize training through additional regularization, CoSub aims to train the sub-branchs by co-training~\cite{zhang2018dml}.
We regard MaskSub as a more generalized framework since CoSub only considers the drop-path method to employ sub-branchs, whereas MaskSub can cover various drop-based techniques, including masking. 

\noindent \textbf{Self-distillation} utilizes supervision from a network itself instead of using a teacher.
ONE~\cite{one} uses a multi-branch ensemble to build superior output for the network and distill ensemble outputs as supervision for each branch.
Some studies~\cite{selfdistill1,multi_exit} utilize the early-exit network for self-distillation. Those studies improve performance by using an entire network as a teacher and an early exit network as a student.
MaskedKD~\cite{son2023maskedkd} utilizes masking to reduce computation for knowledge distillation.
From a self-distillation perspective, MaskSub presents a new insight into constructing the student model (i.e., sub-branch) from the teacher model (i.e., main-branch) utilizing drop-based techniques.
Note that most self-distillation studies are not compatible with recent training recipes~\cite{wightman2021resnet,touvron2022deit3}. Thus, the general applicability of MaskSub is a notable contribution.

\noindent \textbf{Self- and semi-supervised learning} share components with MaskSub.
Contrastive learning incorporates two models with self-distillation loss~\cite{chen2021simsiam,grill2020byol}.
Want et al.~\cite{wang2022tower} introduce a double tower with weak and strong augmentation for each model.
MAE~\cite{he2022masked} uses masked image reconstruction as self-supervision, and supervised MAE~\cite{liang2022supmae} introduces supervised learning as an additional task for MAE. 
MAE and supMAE aim to reconstruct masked images using MAE training recipe, rather than supervised learning. In contrast, MaskSub only relies on label-related loss with a supervised learning recipe.
In semi-supervised learning, UDA~\cite{xie2020uda} introduces a two-branch framework, similar to the main- and sub-branch in MaskSub.
However, MaskSub is more computationally efficient by using masking~\cite{he2022masked} and removing label-consistency checks for unlabeled data.
Also, MaskSub extends the two-branch framework to supervised learning via distillation loss, in contrast to UDA's consistency loss.
While these studies share the fundamental concept with MaskSub and inspired our work, the training techniques for supervised learning differ from those in semi- and self-supervised learning.
Thus, we argue that MaskSub retains its originality and novelty compared to these studies.

\section{Method}

We propose our method Masked Sub-branch (MaskSub) with formulation and pseudo-code in Section~\ref{subsec:augsub}. Section~\ref{subsec:analysis} presents analyses of MaskSub with loss convergence, accuracy, and gradient.
In Section~\ref{subsec:drop_regularizations}, we introduce variants of MaskSub: DropSub, and PathSub. 

\subsection{Masked Sub-branch (MaskSub)}
\label{subsec:augsub}

The cross-entropy loss with the softmax $\sigma(\mathbf{z}) = e^{z_i} / \sum_j e^{z_j}$
for images $\mathbf{x}_i$ and one-hot labels $\mathbf{y}_i  (i \in [1,2, ... , N])$ in a mini-batch with size $N$ is denoted as
\begin{equation}
- \frac{1}{N} \sum_i^N \mathbf{y}_i \mathrm{log} \, (\sigma(f_\theta(\mathbf{x}_i | r_{\mathrm{mask}}=0))),
\label{eq:cross_entropy}
\end{equation}
where $f_\theta$ represents the network used for training. $r_{\mathrm{mask}}$ means a ratio of masked patches in an input image. Since the masking ratio can be easily changed, we denote it as a condition for network function. Based on the value of $r_{\mathrm{mask}}$, certain network features are dropped with probability $r_{\mathrm{mask}}$. Note that we set the default masking ratio to zero for convenience. Then, loss for masking ratio $r \in [0, 1]$ is
\begin{equation}
- \frac{1}{N} \sum_i^N \mathbf{y}_i \mathrm{log} \, (\sigma(f_\theta(\mathbf{x}_i|r_{\mathrm{mask}}=r))).
\label{eq:reg_cross_entropy}
\end{equation}
Typically, a network with mask augmentation is trained with \cref{eq:reg_cross_entropy}.
But, we conjecture that training using \cref{eq:reg_cross_entropy} with a high masking ratio (\ie $r \geq 0.5$) may interfere with loss convergence and induce instability in training.
To ensure training stability, we utilize the model output of equation \cref{eq:cross_entropy}, $f_\theta(\mathbf{x}_i | r_{\mathrm{mask}}=0)$, as guidance for masking augmentation $f_\theta(\mathbf{x}_i | r_{\mathrm{mask}}=r)$ instead of $\mathbf{y}_i$.
In other words, \cref{eq:reg_cross_entropy} is changed as
\begin{equation}
- \frac{1}{N} \sum_i^N \sigma(f_\theta(\mathbf{x}_i| r_{\mathrm{mask}}=0)) \, \mathrm{log} \, (\sigma(f_\theta(\mathbf{x}_i| r_{\mathrm{mask}}=r))).
\label{eq:aux_cross_entropy}
\end{equation}
In our Masked Sub-branch (MaskSub), the average of \cref{eq:cross_entropy} and \cref{eq:aux_cross_entropy} is used as a loss function for the network.
We designate $f_\theta(\mathbf{x}_i | r_{\mathrm{mask}}=0)$ as the main-branch and $f_\theta(\mathbf{x}_i | r_{\mathrm{mask}}=r)$ as the sub-branch. This naming convention is employed because a network with masked inputs appears to be a subset of the entire network.
In \cref{eq:aux_cross_entropy}, the main-branch output $f_\theta(\mathbf{x}_i | r_{\mathrm{mask}}=0)$ is used with stop-gradient. Thus, the sub-branch is trained to mimic the main model, but we want the gradient for the main-branch to be independent of the sub-branch. 
This can be interpreted as self-distillation, where knowledge is transferred from the teacher (main-branch) to the student (sub-branch). 
Note that MaskSub can easily be expanded to binary cross-entropy loss by replacing the softmax function with the sigmoid function, which is used for recent training recipes~\cite{wightman2021resnet,touvron2022deit3}.

Algorithm~\ref{alg:pytorch_pseudocode} describes PyTorch-style pseudo-code of training with MaskSub. The masking ratio is put into the network input. The gradients are calculated on the main and sub-branch average losses. Note that MaskSub does not use any additional data augmentation, optimizer steps, and network parameters for the sub-branch. We use MAE-style random masking~\cite{he2022masked}, removing masked tokens to reduce computation costs by default. It significantly reduces the training cost of MaskSub. Approximately, MaskSub with 50\% masking requires $\times$1.5 computation to standard training. In practical implementation, we separate the main and sub-branch backward passes utilizing the gradient accumulation of PyTorch. So, VRAM for the main-branch can be released before the sub-branch computation, which eliminates the need for additional VRAM for MaskSub training. We will show the impact of MaskSub on diverse cases in Section~\ref{sec:experiment}, including computation analysis in Section~\ref{sec:training_budget}.

\begin{figure}[t]
\centering
\begin{minipage}{1.0\columnwidth}
\vspace{-1em}
\begin{algorithm}[H]
\small
\caption{MaskSub in PyTorch-style pseudo-code}
\label{alg:pytorch_pseudocode}
\begin{lstlisting}[language=Python]
for (x, label) in data_loader:
    o1 = f(x)         # main
    o2 = f(mask(x,r)) # sub (mask ratio: r)
    loss1 = CE(o1, label) / 2
    loss2 = CE(o2, softmax(o1.detach())) / 2
    (loss1+loss2).backward()
    optimizer.step()
\end{lstlisting}
\vspace{-.5em}
\end{algorithm}
\vspace{-2em}
\end{minipage}
\end{figure}

\begin{figure*}[t]
    \centering
    \begin{subfigure}{0.24\textwidth}
        \includegraphics[width=\textwidth]{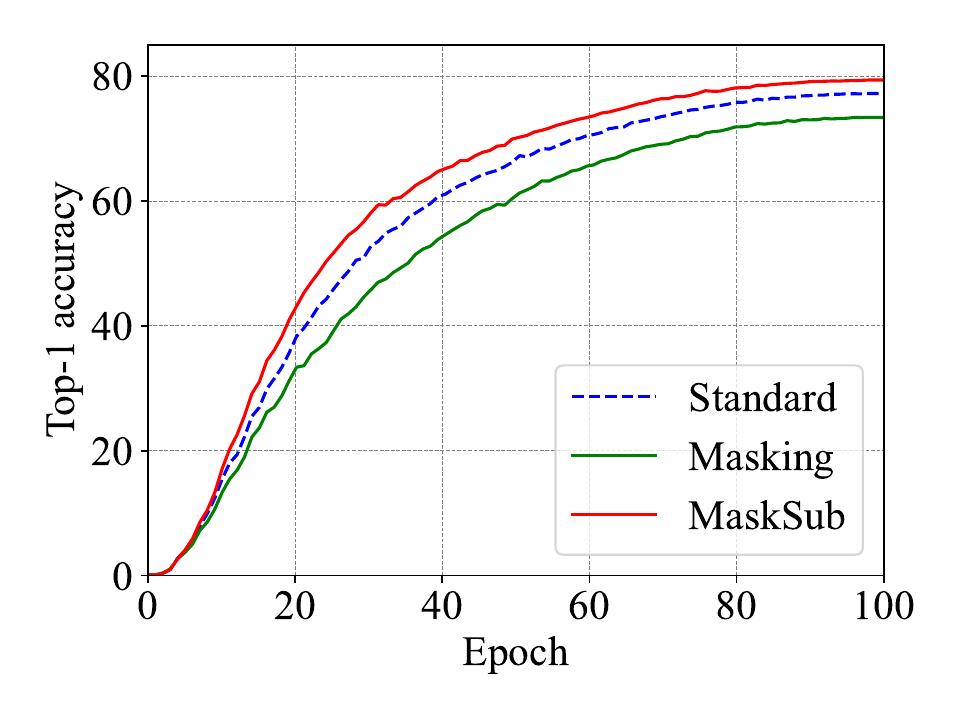}
        \caption{Accuracy}
        \label{fig:analysis_sub1}
    \end{subfigure}
    \hfill
    \begin{subfigure}{0.24\textwidth}
        \includegraphics[width=\textwidth]{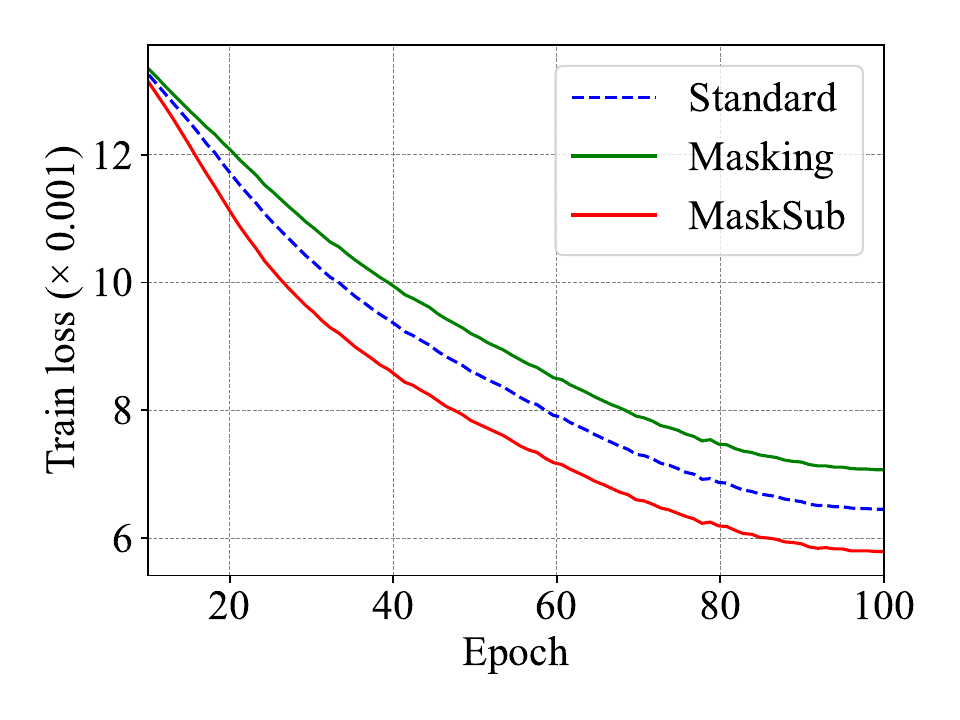}
        \caption{Train loss (standard)}
        \label{fig:analysis_sub2}
    \end{subfigure}
    \hfill
    \begin{subfigure}{0.24\textwidth}
        \includegraphics[width=\textwidth]{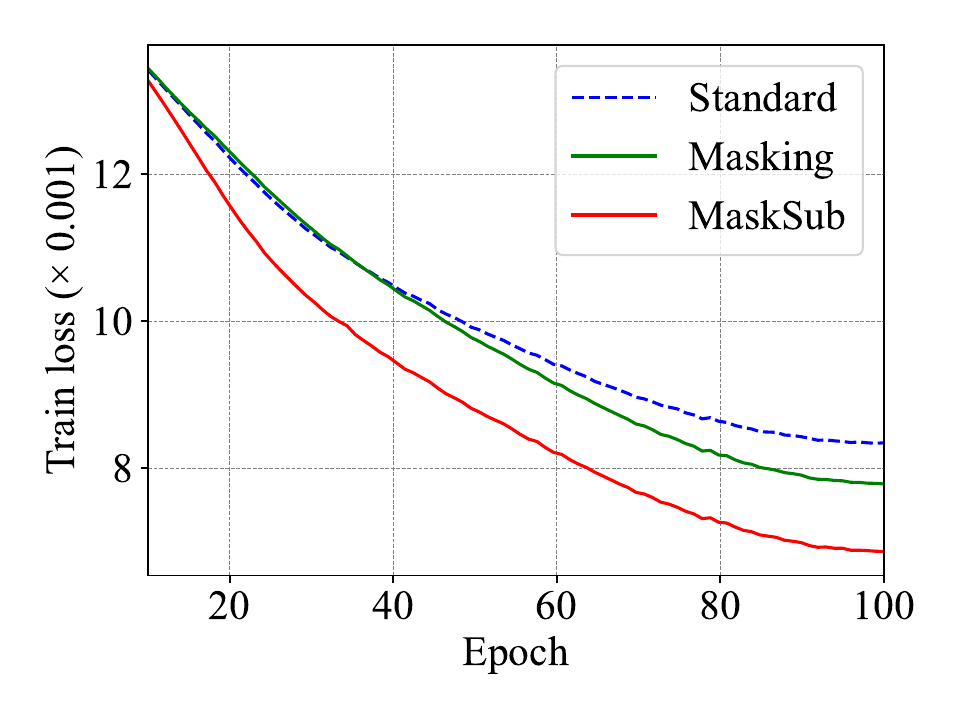}
        \caption{Train loss (masking)}
        \label{fig:analysis_sub3}
    \end{subfigure}
    \hfill
    \begin{subfigure}{0.24\textwidth}
        \includegraphics[width=\textwidth]{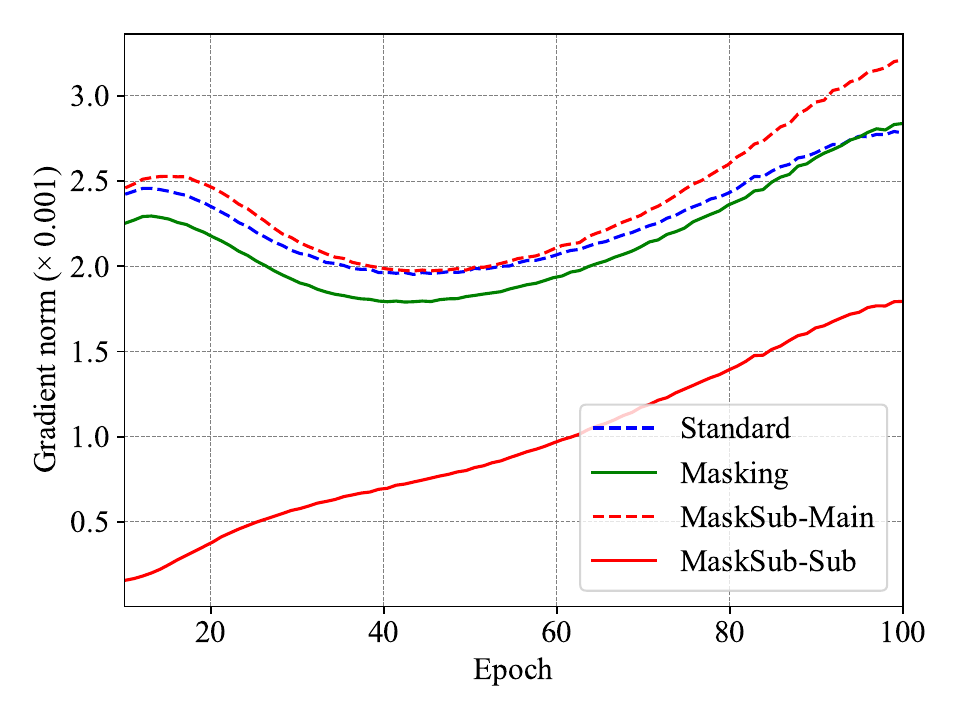}
        \caption{Gradient norm}
        \label{fig:analysis_sub4}
    \end{subfigure}
    \vspace{-.5em}
    \caption{\textbf{MaskSub training analysis}. We use 50\%-random masking to compare three training settings: standard \cref{eq:cross_entropy}, masking \cref{eq:reg_cross_entropy}, and MaskSub. We visualize (a) validation accuracy; (b) train loss without masking; (c) train loss with masking; (d) gradient norm. }
    \label{fig:analysis}
    \vspace{-1em}
\end{figure*}

MaskSub automatically controls the difficulty of the sub-branch.
If the main-branch is close to the ground-truth label, the sub-branch loss aims to attain the ground-truth label under masking. Conversely, if the main-branch fails to converge, the sub-branch loss becomes easy.
This difficulty design is inspired by distillation studies~\cite{mirzadeh2020takd,cho2019effkd,jin2019knowledge}.
The distillation becomes difficult when a high-performance network is used as a teacher~\cite{mirzadeh2020takd,cho2019effkd}.  An early-stage network is easy, and an end-stage network is challenging~\cite{jin2019knowledge}.
Thus, MaskSub can be considered as a sample-wise masking augmentation that is exclusively applied to images that produce successful output in the main-branch.

\subsection{Analysis}
\label{subsec:analysis}

We analyze MaskSub with ViT-B~\cite{dosovitskiy2020vit} for 100 epochs training on ImageNet-1k~\cite{deng2009imagenet}.
Based on DeiT-III~\cite{touvron2022deit3}, we shorten the epoch to 100 epochs and use image resolution $224 \times 224$. 
We compare three settings: standard, masking, and MaskSub.
The standard uses \cref{eq:cross_entropy} as the training loss, and masking augmentation is not used.
For the masking setting, the network is trained with \cref{eq:reg_cross_entropy}. Note that it is a common practice to use a regularization or an augmentation in supervised learning.
We compare those two settings with MaskSub.
For analysis, we measured \cref{eq:cross_entropy} `train loss - standard' and \cref{eq:reg_cross_entropy} `train loss - masking'.
It shows how losses changed by training setting.

Figure~\ref{fig:analysis} shows loss and accuracy trends in random masking 50\% (\ie, $r_{mask} = 0.5$) case.
When random masking is applied to training (\textcolor{ForestGreen}{green}), the masking loss (Figure~\ref{fig:analysis_sub3}) converges better than the standard (\textcolor{blue}{blue}).
However, it significantly degrades the standard train loss (Figure~\ref{fig:analysis_sub2}), resulting in a drop in accuracy (Figure~\ref{fig:analysis_sub1}).
Regularization over a certain strength often causes malicious effects on standard train loss, which decreases accuracy.
As shown in Figure~\ref{fig:analysis_sub2} and \ref{fig:analysis_sub3}, MaskSub improves the loss convergence for both losses, original and masking, which brings an improvement in accuracy.

Figure~\ref{fig:analysis_sub4} explains the learning pattern between main-branch and sub-branch of MaskSub (\cref{eq:aux_cross_entropy}) in the aspect of gradients magnitude for training with random masking 50\%.
The gradient magnitude from the main-branch (MaskSub-Main) is similar to that of other training.
In contrast, gradients from the sub-branch (MaskSub-Sub) have a small magnitude at the early stage. As the learning progresses, the gradients from the sub-branch increase. 
It shows that MaskSub trains the network following our intention: automatic difficulty control.
During the early stage of training, the gradients from the main-branch lead the training. 
Following the progress of the main-branch training, the sub-branch adaptively increases its gradient magnitude and produces a reasonable amount of gradients at the end of training.
In other words, the model training is relaxed from challenging masked inputs at the early stage, while it starts to learn masked input when the original inputs are sufficiently trained.
We claim that the automatic difficulty control of MaskSub could be a general solution to introduce strong augmentation for supervised learning.

\subsection{Expand to drop regularizations}
\label{subsec:drop_regularizations}

We design MaskSub for masking augmentation. 
Due to its simplicity, it can be expanded to drop-based regularizations~\cite{srivastava2014dropout,ghiasi2018dropblock,huang2016stodepth}.
In this section, we introduce two variants of MaskSub: DropSub for dropout~\cite{srivastava2014dropout} and PathSub for drop-path~\cite{huang2016stodepth}.
Since the drop-based regularizations easily adjust their strength by controlling drop probability, MaskSub enables the model to learn dropped features without degrading performance at a standard loss, similar to masking augmentation.
The performance of MaskSub variants is shown in Section~\ref{subsec:drop_exp}. Note that detailed experiments with loss convergence for various drop rates are reported in Table A.5 in the Appendix.

\noindent \textbf{DropSub.}
Dropout~\cite{srivastava2014dropout} is a fundamental drop regularization. Dropout drops random elements of network features with a fixed probability. Since dropout is unrelated to feature structure, every feature element has independent drop probability $p_{drop}$. DropSub is simply implemented by changing $r_{mask}$ to the dropout probability $p_{drop}$. Thus, the sub-branch uses strong dropout, while the main-branch follows a standard training recipe. Due to stability issues, dropout is not preferred in recent training recipes~\cite{touvron2021deit,touvron2022deit3}. However, DropSub enables strong dropout in ViT training and achieves performance improvement.

\noindent \textbf{PathSub.}
Drop-path~\cite{huang2016stodepth,fan2019structuredrop} randomly drops a total feature of the network block with a probability $p_{path}$. 
PathSub is also implemented by changing $r_{mask}$ to the drop-path probability $p_{path}$. Drop-path widely used in training recipes~\cite{touvron2021deit,touvron2022deit3,wightman2021resnet} to adjust the regularization strength~\cite{touvron2022deit3}. Thus, unlike previous cases, the main-branch uses the drop-path following the training recipe, and the sub-branch uses a higher drop probability than the main-branch.

\begin{table}[b]
\centering
\vspace{-1em}
\resizebox{1.0\columnwidth}{!}{
\begin{tabular}{@{}lcccc@{}}
\toprule
\multirow{2}{*}[-0.3em]{Network} & \multicolumn{2}{c}{400 epochs} & \multicolumn{2}{c}{800 epochs} \\ \cmidrule(l){2-3} \cmidrule(l){4-5} 
            & DeiT-III   & \begin{tabular}[c]{@{}c@{}}+ \ours \end{tabular}                   & DeiT-III   & \begin{tabular}[c]{@{}c@{}}+ \ours \end{tabular}              \\ \midrule
ViT-S/16   & 80.4       & \textbf{81.1 \greenpscript{+0.7}} & 81.4       & \textbf{81.7 \greenpscript{+0.3}} \\
ViT-B/16    & 83.5       & \textbf{84.1 \greenpscript{+0.6}} & 83.8       & \textbf{84.2 \greenpscript{+0.4}} \\
ViT-L/16   & 84.5       & \textbf{85.2 \greenpscript{+0.7}} & 84.9       & \textbf{85.3 \greenpscript{+0.4}} \\
ViT-H/14    & 85.1       & \textbf{85.7 \greenpscript{+0.6}} & 85.2       & \textbf{85.7 \greenpscript{+0.5}}                               \\ \bottomrule
\end{tabular}
}
\caption{\textbf{Training from scratch with ViT using the DeiT-III. } MaskSub (50\%) is applied to the ViT training~\cite{touvron2022deit3} on ImageNet-1k.
Note that the training settings are identical to the original ones.
}
\label{table:deit-iii}
\vspace{-1em}
\end{table}

\begin{table}[b]
\centering
\small
\begin{tabular}{@{}llccl@{}}
\toprule
Network & Method & Epochs & Top-1 acc. & Cost\\ \midrule
\multirow{5}{*}[0.1em]{ViT-S} & \textcolor{gray}{DeiT~\cite{touvron2021deit}} & \textcolor{gray}{300} & \textcolor{gray}{79.8} & \textcolor{gray}{-} \\
 & \textcolor{gray}{MAE~\cite{he2022masked}$^\dagger$} & \textcolor{gray}{1600} & \textcolor{gray}{81.4} & \textcolor{gray}{-}  \\
 & DeiT-III~\cite{touvron2022deit3} & 800 & 81.4 & $\times1.0$\\
 & CoSub~\cite{touvron2022co} & 800 & 81.5 & $\times2.0$ \\
 & \cellcolor{lightgreen}\ours & \cellcolor{lightgreen}400 & \cellcolor{lightgreen}81.1 & \cellcolor{lightgreen}\bm{$\times0.75$} \\
 & \cellcolor{lightgreen}\ours & \cellcolor{lightgreen}800 & \cellcolor{lightgreen}\textbf{81.7} & \cellcolor{lightgreen}$\times1.5$ \\ \midrule
\multirow{5}{*}[0.1em]{ViT-B} & \textcolor{gray}{DeiT~\cite{touvron2021deit}} & \textcolor{gray}{300} & \textcolor{gray}{81.8} & \textcolor{gray}{-} \\
 & \textcolor{gray}{MAE~\cite{he2022masked}} & \textcolor{gray}{1600} & \textcolor{gray}{83.6} & \textcolor{gray}{-} \\
 & \textcolor{gray}{SupMAE~\cite{liang2022supmae}} & \textcolor{gray}{400} & \textcolor{gray}{83.6} & \textcolor{gray}{-} \\
 & DeiT-III~\cite{touvron2022deit3} & 800 & 83.8 & $\times1.0$\\
 & CoSub~\cite{touvron2022co} & 800 & \textbf{84.2} & $\times2.0$ \\
 & \cellcolor{lightgreen}\ours & \cellcolor{lightgreen}400 & \cellcolor{lightgreen}84.1 & \cellcolor{lightgreen}\bm{$\times0.75$} \\ 
 & \cellcolor{lightgreen}\ours & \cellcolor{lightgreen}800 & \cellcolor{lightgreen}\textbf{84.2} & \cellcolor{lightgreen}$\times1.5$ \\ \midrule
\multirow{4}{*}[0.1em]{ViT-L} & DeiT-III~\cite{touvron2022deit3} & 800 & 84.9 & $\times1.0$\\
 & CoSub~\cite{touvron2022co} & 800 & \textbf{85.3} & $\times2.0$ \\
 & \cellcolor{lightgreen}\ours & \cellcolor{lightgreen}400 & \cellcolor{lightgreen}85.2 & \cellcolor{lightgreen}\bm{$\times0.75$} \\ 
 & \cellcolor{lightgreen}\ours & \cellcolor{lightgreen}800 & \cellcolor{lightgreen}\textbf{85.3} & \cellcolor{lightgreen}$\times1.5$ \\ \midrule
\multirow{3}{*}[0.1em]{ViT-H} & DeiT-III~\cite{touvron2022deit3} & 800 & 85.2 & $\times1.0$\\
 & CoSub~\cite{touvron2022co} & 800 & \textbf{85.7} & $\times2.0$ \\
 & \cellcolor{lightgreen}\ours & \cellcolor{lightgreen}400 & \cellcolor{lightgreen}\textbf{85.7} & \cellcolor{lightgreen}\bm{$\times0.75$} \\ 
\bottomrule

\end{tabular}
\caption{\textbf{Pre-training methods comparison.} We compare DeiT-III~\cite{touvron2022deit3} + MaskSub with various pre-training methods. MaskSub shows remarkable performances compared to its training cost.
}
\label{table:long_vit}
\vspace{-1em}
\end{table}

\section{Experiments}
\label{sec:experiment}
We validate the effectiveness of our Masked Sub-branch (MaskSub) by applying it to diverse training scenarios.
We claim MaskSub is an easy plug-in solution for various training recipes.
Thus, we strictly follow the original training recipe, including optimizer parameters, learning rate and weight-decay, and regularization parameters.
The only difference between baseline and MaskSub is the masking augmentation for the sub-branch.
We simply set the masking ratio of MaskSub to 50\% across all experiments.
In short, MaskSub does not have a hyper-parameter that varies depending on training scenarios.

\subsection{Training from scratch (pretraining)}

The training recipe in ViTs is a key factor enabling ViT to surpass CNN; thus, the ViT training recipe is an important and active research topic.
We use a state-of-the-art ViT training recipe, DeiT-III~\cite{touvron2022deit3}, as our baseline.
Enhancing DeiT-III by integrating additional techniques is challenging, so we believe improvements made over DeiT-III would represent a new state-of-the-art in ViT training.

ViTs are trained with MaskSub (50\%) on 400 and 800 epochs training. The results are shown in Table~\ref{table:deit-iii}. MaskSub improves performance across all settings. For 400-epochs training, MaskSub improves DeiT-III with substantial margins, which even outperforms 800-epochs trained DeiT-III except for ViT-S/16. MaskSub also demonstrates superior performance when training for 800 epochs. The impact of MaskSub is impressively consistent with larger models like ViT-L/16 and ViT-H/16.
It is worth noting that ViT-H + MaskSub (400 epochs) outperforms ViT-H/16 (800 epochs) with +0.5pp gain, even with half the training epochs.
Thus, MaskSub is an effective way to improve ViT training.

Table~\ref{table:long_vit} shows the performance and computation cost of MaskSub compared to other pretrainings.
In ViT-S and ViT-B, MaskSub outperforms MAE~\cite{he2022masked} with a reasonable performance gap.
Compared to SupMAE~\cite{liang2022supmae}, MaskSup outperforms under the same epochs.
CoSub~\cite{touvron2022co} has comparable performance with MaskSub; however, MaskSub requires less computation costs than CoSub.
Thus, we argue that MaskSub outperforms CoSub.
More comparisons with CoSub are included in Section~\ref{sec:comparison_sub}.

\subsection{Finetuning}

\begin{table}[t]
\centering
\small
\setlength\tabcolsep{6pt}
\begin{tabular}{@{}lcccc@{}}
\toprule
 & Epochs & Network & Baseline & +\ours  \\ \midrule
 \multirow{3}{*}{\begin{tabular}[c]{@{}l@{}}MAE~\cite{he2022masked}\\finetuning \end{tabular}} & 100  & ViT-B/16  &  83.6     & \textbf{83.9 \greenpscript{+0.3}}                   \\
            & 50              & ViT-L/16 & 85.9         & \textbf{86.1 \greenpscript{+0.2}}                   \\
            & 50              & ViT-H/14  & 86.9         & \textbf{87.2 \greenpscript{+0.3}}                   \\ \midrule
 \multirow{2}{*}{\begin{tabular}[c]{@{}l@{}}BEiTv2~\cite{peng2022beit2}\\finetuning\end{tabular}} & 100   & ViT-B/16  & 85.5        &  \textbf{85.6 \greenpscript{+0.1}}   \\
                & 50    & ViT-L/16  & 87.3        &  \textbf{87.4 \greenpscript{+0.1}}  \\ \midrule
\multirow{2}{*}{\begin{tabular}[c]{@{}l@{}}CLIP~\cite{clip}\\finetuning\end{tabular}} & 50  & ViT-B/16  &  84.8     &  \textbf{85.2 \greenpscript{+0.4}}        \\
 & 30 & ViT-L/14 &  87.5        &  \textbf{87.8 \greenpscript{+0.3}}                  \\
\bottomrule
\end{tabular}
\caption{\textbf{ImageNet-1k finetuning.} We report finetuning performance of MAE~\cite{he2022masked}, BEiT v2~\cite{peng2022beit2} and CLIP finetuning~\cite{dong2022ftclip} with MaskSub (50\%). Official weights are used.
}
\label{table:finetune}
\vspace{-1em}
\end{table}

Following the emergence of self-supervised learning~\cite{he2022masked} and visual-language modeling~\cite{clip}, the significance of finetuning has notably increased. Generally, self-supervised learning, such as MAE~\cite{he2022masked} and BEiT~\cite{bao2021beit,peng2022beit2}, does not use supervised labels at pretraining, which makes MaskSub inapplicable for pretraining. However, a standard is to evaluate the model's capability using supervised finetuning after pretraining. Thus, we apply our MaskSub (50\%) to the finetuning stage to verify the effect of MaskSub on finetuning.
Note that we strictly follow the original recipes mentioned below and apply MaskSub (50\%) based on it. All finetuning is conducted using officially released pretrained weights.

We utilize three finetuning recipes: MAE~\cite{he2022masked}, BEiT v2~\cite{peng2022beit2}, and Finetune CLIP~\cite{dong2022ftclip}.
MAE~\cite{he2022masked} is a representative method of masked image models (MIM). 
Since our random masking is motivated by MAE, MaskSub is seamlessly integrated into the MAE finetuning process. 
BEiT v2~\cite{peng2022beit2} utilizes the pretrained CLIP for MIM and achieves superior performance compared to MAE. 
Following the masking strategy of BEiT v2 using mask-token, we adjust MaskSub to masking using mask-token from the pretrained weight instead of MAE-style masking.
Finetune CLIP~\cite{dong2022ftclip} is a finetuning recipe for CLIP~\cite{clip} pretrained weights. MaskSub is applied to finetuning CLIP without change.

\begin{table}[t]
\centering
\small
\setlength\tabcolsep{8pt}
\begin{tabular}{@{}llcc@{}}
\toprule
Network & Method & Top-1 acc. \\ \midrule
\multirow{3}{*}[0.1em]{CLIP-B} & Linear probing~\cite{clip} & 80.2 \\
 & Finetuning~\cite{dong2022ftclip} & 84.8 \\
 & \cellcolor{lightgreen}Finetuning~\cite{dong2022ftclip} + \ours & \cellcolor{lightgreen}\textbf{85.2} \\ \cmidrule{1-3}
\multirow{7}{*}[0.1em]{ViT-B}  & FD-CLIP~\cite{wei2022fdclip} & 84.9 \\
 & MaskDistill~\cite{peng2022unified} & 85.5 \\
 & MVP~\cite{wei2022mvp} & 84.4 \\
 & MILAN~\cite{hou2022milan} & 85.4 \\
 & CAEv2~\cite{zhang2023cae2} & 85.5 \\
 & BEiTv2~\cite{peng2022beit2} & 85.5 \\
 & \cellcolor{lightgreen}BEiTv2~\cite{peng2022beit2} + \ours & \cellcolor{lightgreen}\textbf{85.6} \\
\bottomrule
\end{tabular}
\caption{\textbf{Comparison with CLIP-based training on ImageNet-1k.} Our finetuning experiment is close to the state-of-the-art of ViT-B training. MaskSub applied to BEiTv2~\cite{peng2022beit2} fine-tuning outperforms cutting-edge studies on CLIP-based training.
}
\label{table:long_clip}
\vspace{-1em}
\end{table}

Table~\ref{table:finetune} shows the finetuning results. MaskSub improves the performance of all finetune practices, including large-scale ViT models. 
This is notable as it shows substantial improvement with a short finetuning phase of fewer than 100 epochs compared to the pretraining period of 1600 epochs.
In MAE finetuning, MaskSub improves 0.2 - 0.3pp in all model sizes.
MaskSub is also effective on BEiT v2, which utilizes Relative Position Encoding (RPE)~\cite{liu2021swin,bao2021beit} and block-masking strategy with mask-tokens.
CLIP finetuning also displays that MaskSub achieves substantial improvements. In finetuning CLIP, we report performance at the last epoch rather than selecting the best performance in early epochs.
The best performance of finetuning CLIP with MaskSub is the same as the baseline.
Table~\ref{table:long_clip} demonstrates the impacts of MaskSub compared to cutting-edge CLIP-based training recipes. It shows that MaskSub improves the performance of state-of-the-art training recipes.

\subsection{Hierarchical architecture}
\begin{table}[t]
\centering
\small
\setlength\tabcolsep{4pt}
\begin{tabular}{@{}lcccc@{}}
\toprule
Recipe & Epochs & Network & Baseline & \begin{tabular}[c]{@{}c@{}}+ \ours\end{tabular}  \\ \midrule
 \multirow{3}{*}{\begin{tabular}[c]{@{}c@{}}RSB A2~\cite{wightman2021resnet}\end{tabular}} & \multirow{3}{*}{\begin{tabular}[c]{@{}c@{}}300\end{tabular}} & ResNet50 & 79.7     & \textbf{80.0 \greenpscript{+0.3}} \\
 & & ResNet101 & 81.4         & \textbf{82.1 \greenpscript{+0.7}}                   \\
 & & ResNet152 & 81.8         & \textbf{82.8 \greenpscript{+1.0}}                   \\ \midrule
 \multirow{3}{*}{\begin{tabular}[c]{@{}c@{}}Swin~\cite{liu2021swin}\end{tabular}} & \multirow{3}{*}{\begin{tabular}[c]{@{}c@{}}300\end{tabular}} & Swin-T & 81.3        & \textbf{81.4 \greenpscript{+0.1}}           \\
 & & Swin-S & 83.0     &  \textbf{83.4 \greenpscript{+0.4}} \\
& & Swin-B & 83.5    & \textbf{83.9 \greenpscript{+0.4}}                   \\
\bottomrule
\end{tabular}
\caption{\textbf{ImageNet-1k  with hierarchical architecture.} We show the performance of ResNet~\cite{he2016deep} and Swin Transformer~\cite{liu2021swin} trained from scratch with MaskSub (50\%).}
\label{table:resnet_swin}
\vspace{-1em}
\end{table}

\begin{table}[t]
\centering
\small
\setlength\tabcolsep{8pt}
\begin{tabular}{@{}llcc@{}}
\toprule
Network & Method & Top-1 acc. \\ \midrule
\multirow{6}{*}[0.1em]{ResNet50} & Baseline~\cite{he2016resnet} & 76.1 \\
 & ResNeXt50~\cite{xie2017resnext} + ONE~\cite{one} & 78.2 \\
 & BYOT~\cite{selfdistill1} & 75.2 \\
 & Self-Distillation~\cite{zhang2021self} & 78.3 \\
 & MixSKD~\cite{yang2022mixskd} & 78.8 \\
 & \cellcolor{lightgreen}RSB~\cite{wightman2021resnet} + \ours & \cellcolor{lightgreen}\textbf{80.0} \\
\midrule \multirow{5}{*}[0.1em]{ResNet101} & Baseline~\cite{he2016resnet} & 77.4 \\
 & Self-Distillation~\cite{zhang2021self} & 78.9 \\
 & RSB~\cite{wightman2021resnet} + SD-dropout~\cite{lee2023self} & 81.2 \\
 & RSB~\cite{wightman2021resnet} + PS-KD~\cite{kim2021pskd} & 81.7 \\
 & \cellcolor{lightgreen}RSB~\cite{wightman2021resnet} + \ours & \cellcolor{lightgreen}\textbf{82.1} \\
\midrule \multirow{7}{*}[0.1em]{ResNet152} & Baseline~\cite{he2016resnet} & 78.3 \\
 & PS-KD~\cite{kim2021pskd} & 79.2 \\
 & Self-Distillation~\cite{zhang2021self} & 80.6 \\
 & SD-dropout~\cite{lee2023self} & 75.5 \\
 & RSB~\cite{wightman2021resnet} + SD-dropout~\cite{lee2023self} & 81.8 \\
 & RSB~\cite{wightman2021resnet} + PS-KD~\cite{kim2021pskd} & 82.3 \\
 & \cellcolor{lightgreen}RSB~\cite{wightman2021resnet} + \ours & \cellcolor{lightgreen}\textbf{82.8} \\
\bottomrule
\end{tabular}
\caption{\textbf{Comparison with self-distillation methods.} Based on ResNet, we compare MaskSub with self-distillation methods.
}
\label{table:long_resnet}
\vspace{-1em}
\end{table}

We extend experiments to architectures with hierarchical spatial dimensions: ResNet~\cite{he2016deep} and Swin Transformer~\cite{liu2021swin}.
Unlike ViT, which maintains spatial token length for all layers, those networks change the spatial size of features in the middle of layers, requiring a change in masking strategy.
We apply MaskSub (50\%) to ResNet and Swin Transformer.
We simply fill out masked regions with zero pixels for ResNets and replace masked regions with mask-tokens for Swin Transformer. 
It maintains the spatial structure and enables spatial size reduction of hierarchical architecture.
Following the literature~\cite{woo2023convnext2}, we use random masking with the patch size of $32\times32$.
Note that the computation reduction in MAE-style masking does not apply here; therefore, MaskSub costs double the training budget.
For ResNet, we use an effective training recipe~\cite{wightman2021resnet} with 300 epochs.
The recipe in the original paper~\cite{liu2021swin} is used for the Swin Transformer training.
We strictly follow the training recipes and apply MaskSub without tuning them.

Results are shown in Table~\ref{table:resnet_swin}. 
MaskSub achieves impressive performance gains with ResNet and Swin Transformer as well. 
ResNet and Swin are substantially different architectures from ViT.
Thus, the result implies that the effectiveness of MaskSub is not limited to ViT architectures and is applicable to hierarchical architectures.

\begin{table*}[t]
\centering
\setlength\tabcolsep{6pt}
\vspace{-1mm}
\begin{tabular}{@{}ccccccccc@{}}
\toprule
Model & \begin{tabular}[c]{@{}c@{}}Pretraining\\+ MaskSub \end{tabular} & \begin{tabular}[c]{@{}c@{}}Finetuning\\+ MaskSub \end{tabular} & \begin{tabular}[c]{@{}c@{}}CIFAR100\\\cite{cifar} \end{tabular} & \begin{tabular}[c]{@{}c@{}}CIFAR100\\\cite{cifar} \end{tabular} & \begin{tabular}[c]{@{}c@{}}Flowers\\\cite{flowers} \end{tabular} & \begin{tabular}[c]{@{}c@{}}Cars\\\cite{cars} \end{tabular} & \begin{tabular}[c]{@{}c@{}}iNat-18\\\cite{van2018inaturalist} \end{tabular} & \begin{tabular}[c]{@{}c@{}}iNat-19\\\cite{van2018inaturalist} \end{tabular}\\ \midrule
 \multirow{3}{*}[-0em]{ViT-S/16} & - & - & 98.8 & 90.0 & 94.5 & 80.9 & 70.1 & 76.7 \\
& \textcolor{darkergreen}{\ding{52}} & - & \textbf{98.9} & \textbf{90.6} & 95.2 & 81.2 & 70.8 & 77.0 \\
& \textcolor{darkergreen}{\ding{52}} & \textcolor{darkergreen}{\ding{52}} & 98.8  & 89.9 & \textbf{98.3} & \textbf{92.2} & \textbf{71.2} & \textbf{77.1} \\ \midrule
\multirow{3}{*}[-0em]{ViT-B/16} & - & - & 99.1 & 91.7 & 97.5 & 90.0 & 73.2 & 78.5 \\
& \textcolor{darkergreen}{\ding{52}} & - & \textbf{99.2} & \textbf{91.9} & 97.7 & 90.2 & 73.6 & 78.8 \\
& \textcolor{darkergreen}{\ding{52}} & \textcolor{darkergreen}{\ding{52}} & 98.8 & 89.6 & \textbf{98.7} & \textbf{92.8} & \textbf{73.9} & \textbf{79.1} \\ \bottomrule
\end{tabular}
\caption{\textbf{Transfer learning.} Table shows transfer learning performance with/without MaskSub. We measure the performance when MaskSub is applied to pretraining and finetuning. The standard deviations over three runs are reported in Appendix. %
}
\vspace{-1em}
\label{table:transfer}
\end{table*}

\begin{table}[t]
\centering
\small
\setlength\tabcolsep{4pt}
\begin{tabular}{@{}ccccl@{}}
\toprule
Architecture  & +MaskSub & Epochs  & GPU days  & Accuracy \\ \midrule
\multirow{2}{*}{\begin{tabular}[c]{@{}c@{}}ViT-S/16\end{tabular}} & - & 600 & 22 & 80.7 \\
        & \textcolor{darkergreen}{\ding{52}} & 400 & 22 & \textbf{81.2 \greenpscript{+0.5}} \\ \midrule
\multirow{2}{*}{\begin{tabular}[c]{@{}c@{}}ViT-B/16\end{tabular}} & - & 600 & 26 & 83.7 \\
        & \textcolor{darkergreen}{\ding{52}} & 400 & 25 & \textbf{84.1 \greenpscript{+0.4}} \\ \midrule
\multirow{2}{*}{\begin{tabular}[c]{@{}c@{}}ResNet101\end{tabular}} & -  & 600 & 24 & 81.5 \\
       & \textcolor{darkergreen}{\ding{52}} & 300 & 20 & \textbf{82.1 \greenpscript{+0.6}} \\ \midrule
\multirow{2}{*}{\begin{tabular}[c]{@{}c@{}}ResNet152\end{tabular}} & - & 600 & 32 & 82.0 \\
       & \textcolor{darkergreen}{\ding{52}} & 300 & 29 & \textbf{82.8 \greenpscript{+0.8}} \\ 
\bottomrule
\end{tabular}
\caption{\textbf{Comparison in the same training budget.} Training has been conducted with NVIDIA V100 8 GPUs. GPU days refer to the number of days required for training when using a V100 GPU. }
\label{table:computation_short}
\vspace{-1em}
\end{table}

\begin{table}[t]
    \centering
    \small
    \setlength\tabcolsep{8pt}
    \begin{tabular}{@{}lll@{}}
    \toprule
    Method      & Accuracy & GPU days \\ \midrule
    Baseline~\cite{touvron2022deit3}    & 83.5            & 17.3              \\ \midrule
    DataAug~\cite{hoffer2020repeataug} & 83.5 \greenpscript{+0.0}            & 36.8 \redpscript{+113\%}              \\ 
    GradAug~\cite{yang2020gradaug}     & 83.2 \redpscript{-0.3}           & 39.7 \redpscript{+129\%}              \\
    CoSub~\cite{touvron2022co}       & 83.9 \greenpscript{+0.4}            & 35.3 \redpscript{+104\%}             \\ 
    \ours     & \textbf{84.1\greenpscript{+0.6}}            & \textbf{25.1 \redpscript{+45\%}}             \\ \bottomrule
    \end{tabular}
    \caption{\textbf{ImageNet-1k Comparison.} The table shows performance and computational costs for ViT-B's 400 epoch training.}
    \vspace{-1em}
    \label{table:comparison_deit}
\end{table}

\begin{table}[t]
    \centering
    \small
    \setlength\tabcolsep{5pt}
    \begin{tabular}{@{}rcccc@{}}
    \toprule
     & \multicolumn{2}{c}{Single-scale mIoU} & \multicolumn{2}{c}{Multi-scale mIoU} \\ \cmidrule(l){2-3} \cmidrule(l){4-5}
     & DeiT-III & + MaskSub & DeiT-III & + MaskSub \\ \midrule
    ViT-B & 48.8  & \textbf{49.4 \greenpscript{+0.6}} & 49.7  & \textbf{50.2 \greenpscript{+0.5}}  \\
    ViT-L  & 51.7  & \textbf{52.2 \greenpscript{+0.5}} & 52.3 & \textbf{52.7 \greenpscript{+0.4}}  \\ \bottomrule
    \end{tabular}
    \caption{\textbf{Semantic segmentation on ADE-20k.} UpperNet for ViT backbone is trained with the BEiTv2 segmentation recipe.}
    \vspace{-1em}
    \label{table:segmentation}
\end{table}

\subsection{Self-distillation}
We compare MaskSub with self-distillation methods.
As shown in Table~\ref{table:long_resnet}, most self-distillations report their performance based on weak and old recipes. Thus, they are less effective with cutting-edge recipes (RSB~\cite{wightman2021resnet}) or architectures (ViT).
Otherwise, MaskSub can be plugged into strong training recipes and achieves state-of-the-art with self-distillation loss.
Thus, MaskSub has contributed to practical self-distillation with its broad applicability.

\subsection{Training budget}
\label{sec:training_budget}

We have shown that MaskSub effectively improves the performance of various architectures. 
However, MaskSub requires additional computation costs for the sub-branch, which increases training costs.
Thus, we analyze MaskSub regarding its training costs to determine if MaskSub could be an effective solution within a limited training budget.
We compare MaskSub with training recipes with increased epochs to align with the training budget. 
The training budget is quantified regarding required GPU days when only a single NVIDIA V100 GPU is used for training.
Table~\ref{table:computation_short} shows the results. 
In ViT training, MaskSub outperforms baseline with $\times 1.5$ epochs setting.
Thus, MaskSub is superior to the long epoch training to spend computation costs for training ViT.
For ResNet, we compare 300 epochs MaskSub with 600 epochs training recipe RSB~\cite{wightman2021resnet} A1.
MaskSub outperforms 600 epochs training recipes in ResNet101 and ResNet152.
Consequently, the results show that MaskSub is an effective way to improve training, even considering computation costs for the sub-branch.

\label{sec:comparison_sub}
We compare MaskSub with other training methods: DataAugment~\cite{hoffer2020repeataug}, GradAug~\cite{yang2020gradaug}, and CoSub~\cite{touvron2022co}. DataAugment~\cite{hoffer2020repeataug} uses doubled data augmentations for the same image, which is similar to contrastive learning~\cite{mocov3,caron2021dino}. GradAug~\cite{yang2020gradaug} utilizes a network pruning~\cite{yu2018slimmable} to build sub-network. CoSub introduces a sub-network based on drop-path~\cite{huang2016stodepth} and uses the sub-network as mutual learning~\cite{zhang2018dml}. 
Table~\ref{table:comparison_deit} shows the 400-epochs training from scratch result.
Note that GradAug in Table~\ref{table:comparison_deit} is a 200-epochs training result to adjust computation cost similar to other methods. All augmentation methods require additional computation costs. In particular, GradAug spends almost 300\% of additional training costs compared to original training. On the other hand, our MaskSub only requires a small amount of extra costs (below 50\%), which is a remarkable advantage in training. With the smallest computation, our MaskSub achieves substantial performance improvements. MaskSub performs superior to CoSub in all cases.%

\begin{table*}[t]
\centering
\setlength\tabcolsep{8pt}
\begin{tabular}{@{}ccccccccccc@{}}
\toprule
Model & +MaskSub & MNLI & QQP & QNLI & SST-2 & CoLA & STS-B & MRPC & RTE & Average \\ \midrule
\multirow{2}{*}[-0em]{\begin{tabular}[c]{@{}c@{}}BERT~\cite{devlin2018bert}\\ \textit{base} \end{tabular}} & - & 84.1 & 87.5 & 91.0 & 91.6 & 54.7 & \textbf{87.0} & 88.5 & 62.8 & 80.9 \\
 & \textcolor{darkergreen}{\ding{52}} & \textbf{84.5} & \textbf{87.7} & \textbf{91.3} & \textbf{91.9} & \textbf{58.3} & 86.8 & \textbf{89.2} & \textbf{63.2}  & \textbf{81.6} \\ \midrule
\multirow{2}{*}[-0em]{\begin{tabular}[c]{@{}c@{}}BERT~\cite{devlin2018bert}\\ \textit{large} \end{tabular}}  & - & 86.8 & 88.2 & 92.3 & 93.8 & 63.3 & \textbf{89.3} & \textbf{92.0} & \textbf{69.7} & 84.4 \\
 & \textcolor{darkergreen}{\ding{52}} & \textbf{87.1} & \textbf{89.0} & \textbf{92.7} & \textbf{94.0} & \textbf{65.2} & 88.6 & 91.5 & 69.3 & \textbf{84.7} \\
\bottomrule
\end{tabular}
\caption{\textbf{GLUE~\cite{wang2018glue} benchmark with BERT~\cite{devlin2018bert}.} We apply MaskSub on GLUE benchmark to validate the effect of MaskSub on language model fine-tuning. MaskSub effectively improves BERT finetuning performance.}
\vspace{-1em}
\label{table:glue}

\end{table*}

\begin{table}[t]
    \centering
    \setlength\tabcolsep{8pt}
    \begin{tabular}{@{}ll@{}}
    \toprule
    Training method & \begin{tabular}[c]{@{}l@{}}ImageNet-1k\\ Zero-shot acc. \end{tabular} \\ \midrule
    CLIP~\cite{clip} & 33.5 \\ 
    CLIP~\cite{clip} + Masking & 29.8 \redpscript{-3.7} \\
    CLIP~\cite{clip} + MaskSub & \textbf{37.6 \greenpscript{+4.1}} \\
    \bottomrule
    \end{tabular}
    \caption{\textbf{MaskSub on CLIP pretraining} with ViT-B/32. We apply MaskSub to CLIP, vision and language, pre-training process. MaskSub is effective for CLIP pre-training.
    }
    \label{table:clip}
\vspace{-1em}
\end{table}

\subsection{Transfer learning}

Improvement in pretraining can boost the performance of downstream tasks~\cite{kornblith2019better}. We measure the transfer learning performance of MaskSub using 800 epochs pretrained weight from Table~\ref{table:deit-iii}. CIFAR-10~\cite{cifar}, CIFAR-100~\cite{cifar}, Oxford Flowers-102~\cite{flowers}, Stanford Cars~\cite{cars} and iNaturalist~\cite{van2018inaturalist} are used for finetuning datasets. We use the AdamW training recipe~\cite{touvron2022deit3} and also evaluate performance when MaskSub (50\%) is applied to the finetuning process. Table~\ref{table:transfer} shows the results. The backbone pretrained with MaskSub consistently outperforms the DeiT-III backbone across all cases. Moreover, when MaskSub is applied to the finetuning, it further boosts performance except CIFAR~\cite{cifar}.

We verify transfer learning to semantic segmentation task on ADE-20k~\cite{zhou2017ade1}. We train UperNet~\cite{xiao2018uppernet} training recipe~\cite{peng2022beit2} and utilize pretrained weight from Table~\ref{table:deit-iii}. Table~\ref{table:segmentation} shows the segmentation results of single-scale and multi-scale
evaluations. On both evaluations, the backbone pretrained with MaskSub demonstrates superior performance, consistent for
ViT-B and ViT-L.

\subsection{Beyond vision domain}

MaskSub can be extended to domains beyond images, as long as the masking is applicable.
Thus, we apply MaskSub to two additional tasks beyond the image domain: GLUE~\cite{wang2018glue} benchmark and CLIP~\cite{clip} pretraining. The first task is a text-classification benchmark GLUE~\cite{wang2018glue}. We use BERT~\cite{devlin2018bert} as a pretrained model and apply MaskSub with 15\% masking following the masking ratio of BERT. As shown in Table~\ref{table:glue}, MaskSub improves text-classification performance. 
MaskSub is also applied to CLIP~\cite{clip} pretraining. Table~\ref{table:clip} shows the results. CLIP trained with MaskSub (50\%) shows improved zero-shot performance. 
Experimental details are in Appendix. %
These results verify that MaskSub has remarkable impacts not only on the vision but also on the language and vision\&language domain.

\begin{table}[t]
    \centering
    \setlength\tabcolsep{4pt}
    \begin{tabular}{@{}cccccc@{}}
    \toprule
    Architecture & Baseline & \ours & DropSub & PathSub  \\ \midrule
    ViT-S/16   &   80.4 & \textbf{81.1 \greenpscript{+0.7}} & 80.6 \greenpscript{+0.2} & 80.8 \greenpscript{+0.4} \\
    ViT-B/16    &   83.5 & \textbf{84.1 \greenpscript{+0.6}} & 83.8 \greenpscript{+0.3} & 83.8 \greenpscript{+0.3} \\ \midrule
    Computation &  $\times 1.0$  &  $\times 1.5$ &  $\times 2.0$ &  $\times 2.0$ \\ \bottomrule
    \end{tabular}
    \caption{\textbf{Comparison of MaskSub variants.} We validate drop-based variants of MaskSub. The sub-branch training improves performance with other drop-methods. But, MaskSub shows the best improvement with the smallest computations.}
    \label{table:deit-iii_drop}
    \vspace{-1em}
\end{table}

\subsection{Extending to drop regularizations}
\label{subsec:drop_exp}

In Section~\ref{subsec:drop_regularizations}, we expand MaskSub with drop regularizations~\cite{srivastava2014dropout,huang2016stodepth}.
We validate the performance of MaskSub variants on a 400 epochs training with Deit-III. 
We use masking~\cite{he2022masked} (50\%), dropout~\cite{srivastava2014dropout} (0.2), and drop-path~\cite{huang2016stodepth} (baseline + 0.1) for MaskSub, DropSub, and PathSub, respectively. Table~\ref{table:deit-iii_drop} shows the results.
Variants of MaskSub outperform the baseline. Among the three, MaskSub shows the best performance. 
Also, MaskSub has the lowest computation costs due to MAE~\cite{he2022masked}-style computation reduction. Thus, we conclude that MaskSub (50\%) is the best in practice compared to variants with drop regularizations.
Note that Table A.5 in Appendix includes more results. 

\section{Conclusion}

In this work, we have presented a new way to introduce masking augmentation to supervised learning. Our method, Masked Sub-branch (MaskSub), is designed to leverage masking augmentation within a sub-branch, which is separated from main training and uses a relaxed loss function.
Our extensive analysis reveals that MaskSub effectively mitigates malicious effects of heavy masking while accelerating the convergence, yielding superior performance. 
We verify MaskSub on various training recipes with diverse architecture. 
Notably, MaskSub demonstrates impressive performance improvements across various scenarios. 
We claim that MaskSub is a substantial advancement in training recipes and contributes to using augmentations.

{
    \small
    \bibliographystyle{ieeenat_fullname}
    \bibliography{main}
}

\clearpage
\onecolumn
\section*{Appendix}
\appendix
\numberwithin{equation}{section}
\numberwithin{figure}{section}
\numberwithin{table}{section}

\section{Experiments (cont'd)}

\subsection{Training budget}
\begin{figure}[h]
  \vspace{-1em}
    \centering
    \begin{subfigure}{0.48\columnwidth}
        \includegraphics[width=\textwidth]{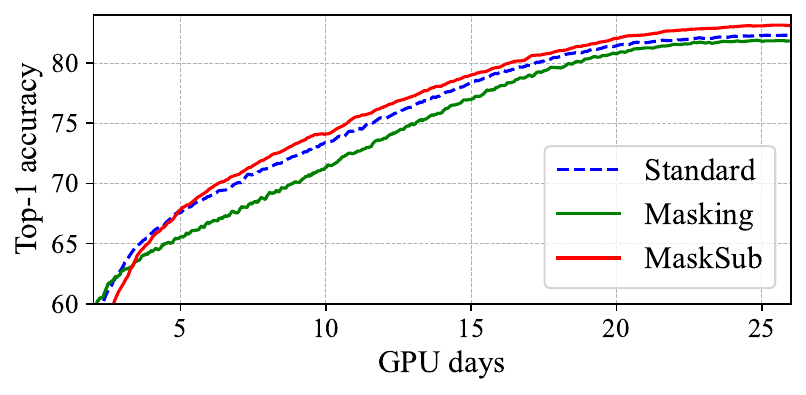}
        \caption{Accuracy}
    \end{subfigure}
    \hfill
    \begin{subfigure}{0.48\columnwidth}
        \includegraphics[width=\textwidth]{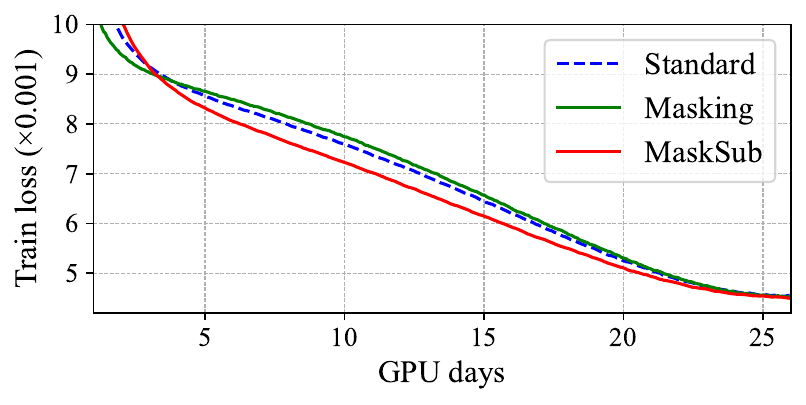}
        \caption{Train loss (standard)}
    \end{subfigure}
    \hfill
    \begin{subfigure}{0.48\columnwidth}
        \includegraphics[width=\textwidth]{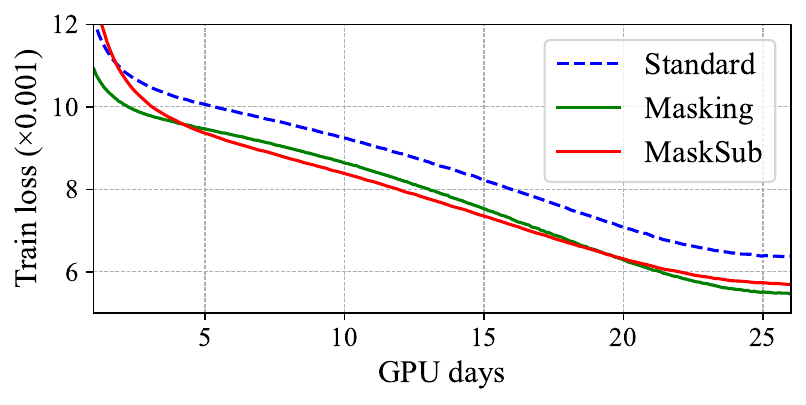}
        \caption{Train loss (masking)}
    \end{subfigure}
    \hfill
    \begin{subfigure}{0.48\columnwidth}
        \includegraphics[width=\textwidth]{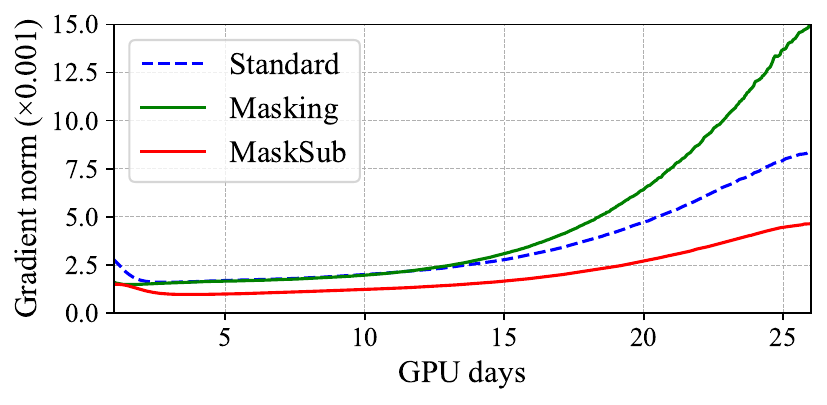}
        \caption{Gradient norm}
    \end{subfigure}
\caption{\textbf{MaskSub training analysis for training budget.} The figures show training analysis experiments in Fig. \textcolor{teal}{2} with the training budget (GPU days) as x-axis. Even considering its additional training budgets, MaskSub effectively improves convergence and accuracy.}
\label{fig:analysis_budget}
\end{figure}

We report an extensive comparison of the training budget in addition to Table~\ref{table:long_vit} and \ref{table:computation_short}. %
Table~\ref{table:computation} shows the performance improvements of MaskSub compared to training recipes with increased epochs to have the same training costs as MaskSub.
In most cases, MaskSub outperforms its counterparts even with the same training computation costs.
The results imply that MaskSub is a better solution than increasing training epochs to improve model performance. 
Note that the performance of MAE finetuning is lower than that of the original epoch since it utilizes the benefits of early stop training to prevent overfitting. Table~\ref{table:comparison_finetune} presents a comparison with other training methods using additional augmentations in MAE~\cite{he2022masked} fine-tuning task. Similar to Table~\ref{table:computation_short}, MaskSub shows the impressive trade-off between performance and computation costs, even in the fine-tuning task. Fig.~\ref{fig:analysis_budget} shows training analysis in Fig.~\ref{fig:analysis} with the same training budget setting. We use GPU days as an x-axis of analysis. MaskSub is effective for loss convergence and accuracy, even considering the additional training budget.
\begin{table*}[h]
\centering
\small
\setlength\tabcolsep{4pt}
\caption{\textbf{Comparison under the equivalent training budget.} All trainings are conducted on NVIDIA V100 8 GPUs. GPU days refer to the number of days required for training when using a single V100 GPU. }
\begin{tabular}{@{}lccccccccc@{}}
\toprule
      & Architecture  & \begin{tabular}[c]{@{}c@{}}Training recipe\end{tabular} & +MaskSub & Epochs  & GPU days  & Accuracy \\ \midrule
\multirow{8}{*}[-1em]{\begin{tabular}[c]{@{}l@{}}DeiT-III\\Training\end{tabular}} & \multirow{4}{*}{\begin{tabular}[c]{@{}c@{}}ViT-S/16\end{tabular}} & \multirow{2}{*}{DeiT-III} & - & 600 & 22 & 80.7 \\
       &         &  & \textcolor{darkergreen}{\ding{52}} & 400 & 22 & \textbf{81.2 \greenpscript{+0.5}} \\ \cmidrule{3-7}
       &         & \multirow{2}{*}{DeiT-III} & - & 1200 & 45  & 81.6 \\ 
       &         &  & \textcolor{darkergreen}{\ding{52}} & 800 & 44 & \textbf{81.7 \greenpscript{+0.1}} \\ \cmidrule{2-7}
       & \multirow{4}{*}{\begin{tabular}[c]{@{}c@{}}ViT-B/16\end{tabular}} & \multirow{2}{*}{DeiT-III} & - & 600 & 26 & 83.7 \\
       &         &  & \textcolor{darkergreen}{\ding{52}} & 400 & 25 & \textbf{84.1 \greenpscript{+0.4}} \\ \cmidrule{3-7}
       &         & \multirow{2}{*}{DeiT-III} & - & 1200 &  52  & 83.8 \\ 
       &         &  & \textcolor{darkergreen}{\ding{52}} & 800 & 50 & \textbf{84.2 \greenpscript{+0.4}} \\ \midrule
\multirow{4}{*}[-0.4em]{\begin{tabular}[c]{@{}l@{}}MAE\\ Finetuning\end{tabular}} & \multirow{2}{*}{\begin{tabular}[c]{@{}c@{}}ViT-B/16\end{tabular}} & \multirow{2}{*}{MAE Finetune} & - & 150 & 9 & 83.5 \\
       &         &  & \textcolor{darkergreen}{\ding{52}} & 100 & 9 & \textbf{83.9 \greenpscript{+0.4}} \\ \cmidrule{2-7}
       & \multirow{2}{*}{\begin{tabular}[c]{@{}c@{}}ViT-L/16\end{tabular}} & \multirow{2}{*}{MAE Finetune} & - & 75 & 14 & 85.5 \\
       &         &  & \textcolor{darkergreen}{\ding{52}} & 50 & 14 & \textbf{86.1 \greenpscript{+0.6}} \\ \midrule
\multirow{6}{*}[-0.5em]{\begin{tabular}[c]{@{}l@{}}ResNet\\Training\end{tabular}} & \multirow{2}{*}{\begin{tabular}[c]{@{}c@{}}ResNet50\end{tabular}} & RSB A1 & - & 600 & 22 & 80.4 \\
       &         & RSB A2 & \textcolor{darkergreen}{\ding{52}} & 300 & 14 & \textbf{80.0 \redpscript{-0.4}} \\ \cmidrule{2-7}
       & \multirow{2}{*}{\begin{tabular}[c]{@{}c@{}}ResNet101\end{tabular}} & RSB A1 & -  & 600 & 24 & 81.5 \\
       &         & RSB A2 & \textcolor{darkergreen}{\ding{52}} & 300 & 20 & \textbf{82.1 \greenpscript{+0.6}} \\ \cmidrule{2-7}
       & \multirow{2}{*}{\begin{tabular}[c]{@{}c@{}}ResNet152\end{tabular}} & RSB A1 & - & 600 & 32 & 82.0 \\
       &         & RSB A2 & \textcolor{darkergreen}{\ding{52}} & 300 & 29 & \textbf{82.8 \greenpscript{+0.8}} \\ 
\bottomrule
\end{tabular}
\label{table:computation}
\vspace{-1em}
\end{table*}

\clearpage

\begin{table}[h]
    \centering
    \small
    \setlength\tabcolsep{4pt}
    \caption{\textbf{MAE finetuning comparison.} The table displays performances and GPU costs for MAE ViT-B fine-tuning.}
    \begin{tabular}{@{}lll@{}}
    \toprule
    Method      & Accuracy & GPU days \\ \midrule
    Baseline~\cite{touvron2022deit3}    & 83.6            & 6.0              \\ \midrule
    DataAug~\cite{hoffer2020repeataug} & 83.1 \redpscript{-0.5}            & 11.8 \redpscript{+97\%}              \\ 
    GradAug~\cite{yang2020gradaug}     & \textbf{84.0\greenpscript{+0.4}}           & 23.4 \redpscript{+290\%}              \\
    CoSub~\cite{touvron2022co}       & 83.8 \greenpscript{+0.2}            & 11.0 \redpscript{+83\%}             \\ 
    \ours  & 83.9 \greenpscript{+0.3}            & \textbf{8.6 \redpscript{+43\%}}             \\ \bottomrule
    \end{tabular}
    \label{table:comparison_finetune}
\end{table}

\label{sec:training_budget}

\subsection{Downstream tasks}

\noindent \textbf{Transfer learning} performances are presented in Table~\ref{table:transfer} in the main paper.

\noindent \textbf{Semantic segmentation} results are included in Table~\ref{table:segmentation} in the main paper.

\noindent \textbf{Object detection and instance segmentation.} We utilize Cascaded Mask R-CNN~\cite{cai2019cascade} with the ViT backbones~\cite{li2022vitdet} for MS COCO~\cite{lin2014mscoco}, which conducts object detection and instance segmentation simultaneously. ViTDet~\cite{li2022vitdet} is used as a training recipe for this experiment. Table~\ref{table:detection} shows the results. The metric $AP^{box}$ quantifies the performance in object detection, while $AP^{mask}$ provides performance in instance segmentation. In both measures, the backbone pretrained with MaskSub outperforms the DeiT-III backbone.

\begin{table}[h]
    \centering
    \small
    \setlength\tabcolsep{5pt}
    \caption{\textbf{Detection and instance segmentation on MS COCO.} Cascaded Mask R-CNN with ViT-B is used.}
    \begin{tabular}{@{}rcc@{}}
    \toprule
     & $AP^{box}$ & $AP^{mask}$ \\ \midrule
    DeiT-III & 50.7  & 43.6   \\
    +MaskSub  & \textbf{50.9 \greenpscript{+0.2}}  & \textbf{43.9 \greenpscript{+0.3}} \\ \bottomrule
    \end{tabular}
    \label{table:detection}
\vspace{-1em}
\end{table}

\subsection{Robustness}

We evaluate the impact of MaskSub in various robustness benchmarks.
We use models trained for 800 epochs in Table~\ref{table:deit-iii}. %
Table~\ref{table:robustness} shows the results.
ViT models trained with MaskSub demonstrate superior performance in in-distribution metrics and all out-of-distribution metrics.
Specifically, MaskSub outperforms each baseline in natural adversarial examples (IN-A~\cite{hendrycks2021natural}), objects in different styles and textures (IN-R~\cite{hendrycks2021many}), controls in rotation, background, and viewpoints (ObjNet~\cite{barbu2019objectnet}), and detecting spurious correlations with background~\cite{djolonga2021robustness} (SI-size, SI-loc, and SI-rot).
The results demonstrate that the improvement of MaskSub is not limited to the ImageNet-1k validation and has been verified across various robustness metrics.

\begin{table*}[h!]
\centering
\small
\setlength\tabcolsep{4pt}
\vspace{-1mm}
\caption{\textbf{Robustness benchmark.} Table shows the robustness benchmark for ViT pretrained with/without MaskSub. $\uparrow$ means higher score is better robustness, while $\downarrow$ indicates lower score is better. %
}
\begin{tabular}{@{}rccccccccccc@{}}
\toprule
Model & +MaskSub & IN-1k($\uparrow$) & IN-V2($\uparrow$) & IN-Real($\uparrow$) & IN-A($\uparrow$) & IN-R($\uparrow$) & IN-C($\downarrow$) & ObjNet($\uparrow$) & SI-size($\uparrow$) & SI-loc($\uparrow$) & SI-rot($\uparrow$) \\ \midrule
\multirow{2}{*}{ViT-S} & - & 81.4 & 70.1 & 87.0 & 23.4 & 46.4 & 44.8 & 32.6 & 55.0 & 39.8 & 37.8 \\
 & \textcolor{darkergreen}{\ding{52}} & \textbf{81.7} & \textbf{71.0} & \textbf{87.4} & \textbf{26.9} & \textbf{47.2} & \textbf{43.6} & \textbf{33.5} & \textbf{56.7} & \textbf{42.5} & \textbf{39.9} \\ \midrule
\multirow{2}{*}{ViT-B} & - & 83.8 & 73.4 & 88.2 & 36.8 & 54.1 & 38.1 & 35.7 & 58.0 & 42.7 & 41.5 \\
 & \textcolor{darkergreen}{\ding{52}} & \textbf{84.2} & \textbf{74.0} & \textbf{88.6} & \textbf{41.9} & \textbf{54.4} & \textbf{36.9} & \textbf{37.2} & \textbf{59.0} & \textbf{44.8} & \textbf{43.3} \\ \midrule
\multirow{2}{*}{ViT-L} & - & 84.9 & 74.8 & 88.8 & 45.3 & 57.4 & 33.9 & 38.8 & 59.8 & 46.5 & 45.0 \\
 & \textcolor{darkergreen}{\ding{52}} & \textbf{85.3} & \textbf{75.8} & \textbf{89.2} & \textbf{51.1} & \textbf{58.5} & \textbf{32.4} & \textbf{40.0} & \textbf{60.2} & \textbf{46.8} & \textbf{45.9} \\ \midrule
\multirow{2}{*}{ViT-H} & - & 85.2 & 75.7 & 89.2 & 51.9 & 58.8 & 32.8 & 40.1 & 61.9 & 49.0 & 46.8 \\
 & \textcolor{darkergreen}{\ding{52}} & \textbf{85.7} & \textbf{76.5} & \textbf{89.6} & \textbf{58.3} & \textbf{59.9} & \textbf{31.2} & \textbf{41.7} & \textbf{62.4} & \textbf{50.1} & \textbf{48.4} \\

\bottomrule
\end{tabular}
\label{table:robustness}
\end{table*}

\clearpage

\subsection{Extending to drop regularizations}
\label{asubsec:drop_regularize}

To generalize MaskSub to drop regularizations, we demonstrate MaskSub with drop-out~\cite{srivastava2014dropout} (DropSub) and drop-path~\cite{huang2016stodepth} (PathSub) in Table~\ref{table:deit-iii_drop}. %
In this section, we present additional analysis to verify MaskSub variants. The experiment setup is the same as in Section 3.2. %
The results are shown in Table~\ref{table:analysis}.
We validate three MaskSub methods for three kinds of regularizations: random masking~\cite{he2022masked}, drop-out~\cite{srivastava2014dropout}, and drop-path~\cite{huang2016stodepth}. ``Original" means the original training recipe trained with Eq.~\ref{eq:cross_entropy},
and none of the drop regularizations is used. ``Single model" shows the performance when the network is trained with Eq.~\ref{eq:reg_cross_entropy}. 
We compare those common practices with ``MaskSub variants", models trained with Eq.~\ref{eq:cross_entropy} 
and \ref{eq:aux_cross_entropy}.
For analysis, we measured Eq.~\ref{eq:cross_entropy} 
``train loss (original)" and Eq.~\ref{eq:reg_cross_entropy} 
``train loss (drop)" regardless of the loss used for training. It shows how losses changed by training setting.
The results demonstrate that MaskSub improves training in all three cases.
In all cases, MaskSub improves original and drop loss convergence, which is connected to superior accuracy compared to original training.
\begin{table*}[h]
\centering
\small
\setlength\tabcolsep{5pt}
\caption{\textbf{Analysis with drop regularizations.} This table shows 100 epochs of the ViT-B performance trained with drop regularizations. Note that training loss scale $10^{-3}$ is omitted for simplicity. The table presents the average values over three separate runs, and the standard deviations are reported in Table~\ref{atable:analysis}. We observe that MaskSub consistently enhances the baselines significantly and also decreases the training loss at the main branch. This suggests that our method effectively facilatates training with regularizations.}
\begin{tabular}{@{}rccccccc@{}}
\toprule
         & & \multicolumn{3}{c}{Single model}  & \multicolumn{3}{c}{MaskSub variants} \\ \cmidrule(l){3-5} \cmidrule(l){6-8}
         & \begin{tabular}[c]{@{}c@{}}Drop\\ ratio\end{tabular} & Accuracy & \begin{tabular}[c]{@{}c@{}}Train loss\\(original)\end{tabular} & \begin{tabular}[c]{@{}c@{}}Train loss\\ (drop)\end{tabular} & Accuracy & \begin{tabular}[c]{@{}c@{}}Train loss\\ (original) \end{tabular} & \begin{tabular}[c]{@{}c@{}}Train loss\\ (drop)\end{tabular} \\ \cmidrule(l){1-2} \cmidrule(l){3-5} \cmidrule(l){6-8}
Original & - & 77.4  &  6.42 & -    &   -       & -   & -   \\ \cmidrule(r){1-2} \cmidrule(l){3-5} \cmidrule(l){6-8}
\multirow{3}{*}{Masking~\cite{he2022masked}} & 25\% & 76.3  &  6.60 & 6.96   & 79.0  & 5.89  & 6.38  \\ 
 & 50\% & 73.8  &  7.02 & 7.77   & 79.4  & 5.81  & 6.89  \\
 & 75\% & 67.3  &  8.08 & 9.27   & 79.2  & 5.84  & 8.15  \\ \cmidrule(r){1-2} \cmidrule(l){3-5} \cmidrule(l){6-8}
\multirow{3}{*}{Drop-out~\cite{srivastava2014dropout}} & 0.1 & 76.1  &  6.60 & 6.87  & 79.1   & 5.88  & 6.32  \\ 
 & 0.2 & 74.1  &  6.95 & 7.34  & 79.1   & 5.82  & 6.57  \\
 & 0.3 & 71.6  &  8.34 & 7.79  & 79.1   & 5.84  & 6.90  \\ \cmidrule(r){1-2} \cmidrule(l){3-5} \cmidrule(l){6-8}
\multirow{3}{*}{Drop-path~\cite{huang2016stodepth}} & 0.1 & 77.4  &  6.42 & 6.42  & 78.4  & 6.11 & 6.11   \\ 
& 0.2 & 74.9  &  6.74 & 7.19  & 78.7   & 5.91   & 6.48   \\ 
 & 0.3 & 71.6  &  7.31 & 8.04  & 78.8   & 5.87   & 7.02   \\  \bottomrule
\end{tabular}
\label{table:analysis}
\vspace{-1em}
\end{table*}

\clearpage
\subsection{Ablation studies}

We conduct ablation studies of MaskSub using DeiT-III training settings to validate the role of KD loss and the effects of hyperparameters. Table~\ref{atable:wokd} shows the impacts of KD loss in MaskSub. Performance improvement by MaskSub is significantly limited when KD loss is not used. Table~\ref{atable:mask_ratio} and Table~\ref{atable:loss_weights} show analysis for masking ratio and loss weights,  respectively. Note that we train for 100 epochs with the DeiT-III setting, and the baseline without MaskSub is 77.4. In Table~\ref{atable:multi_photo}, we test additional ways to improve MaskSub's performance: multiple sub-branch and photometric augmentation on sub-branch. We train ViT-B/16 with 400 epochs DeiT-III setting with MaskSub variants. Single sub-branch shows the best performance with the smallest computation budget. Color-jittering improves MaskSub with negligible cost increases, but Gaussian blur is ineffective.

\begin{table}[h]
    \small
    \centering
\setlength\tabcolsep{5pt}
    \caption{\small \textbf{Effect of KD loss.} We analyze removing KD loss of the sub-model. We use the DeiT-III 400 epochs setting as in Table~\ref{table:deit-iii}.
    }
    \begin{tabular}{@{}lccc@{}}
    \toprule
          & \multirow{2}{*}{Baseline} & \multicolumn{2}{c}{MaskSub} \\ \cmidrule(l){3-4} 
          &                           & w/o KD loss  & with KD loss \\ \midrule
    ViT-S/16 & 80.4  & 80.5 \greenpscript{+0.1} & \textbf{81.1 \greenpscript{+0.7}}      \\ 
    ViT-B/16 & 83.5  & 83.6 \greenpscript{+0.1} & \textbf{84.1 \greenpscript{+0.6}}      \\ \bottomrule
    \end{tabular}
    \label{atable:wokd}
    \vspace{-0.4cm}
\end{table}

\begin{table}[h]
    \small
    \centering
    \setlength\tabcolsep{5pt}
    \caption{\small \textbf{Mask ratio.} Table shows MaskSub with various mask ratios using DeiT-III ViT-B/16 100 epochs setting. 
    }
    \begin{tabular}{@{}lcccccc@{}}
    \toprule
    Mask ratio & 0\% & 25\% & 40\% & 50\% & 60\% & 75\% \\ \midrule
    Accuracy   & 78.4  & 79.0 & 79.2 & 79.2 & 79.3  & 79.2 \\
    Costs  &  $\times 2.0$   & $\times 1.75$ & $\times 1.6$ & $\times 1.5$ & $\times 1.4$  & $\times 1.25$ \\ 
    \bottomrule
    \end{tabular}
    
    \label{atable:mask_ratio}
\end{table}

\begin{table}[h]
    \small
    \centering
    \tabcolsep=0.4em
    \setlength\tabcolsep{5pt}
    \caption{\textbf{Loss weights.} Different loss weights are compared using ViT-B/16 with 100 epochs setting. Note that MaskSub uses ($\frac{1}{2}$, $\frac{1}{2}$).}
    \begin{tabular}{@{}lcccccc@{}}
    \toprule
    \begin{tabular}[c]{@{}c@{}}Weights (CE, KD)\end{tabular} & (1, 0) & ($\frac{3}{4}$, $\frac{1}{4}$) & ($\frac{2}{3}$, $\frac{1}{3}$) & ($\frac{1}{2}$, $\frac{1}{2}$) & ($\frac{1}{3}$, $\frac{2}{3}$) & ($\frac{1}{4}$, $\frac{3}{4}$)   \\ \midrule
    Accuracy       & 77.4    & 78.5 &  78.7  &  79.2  & 79.2 & 78.5 \\ 
    \bottomrule
    \end{tabular}
    \label{atable:loss_weights}
\end{table}

\begin{table}[h]
    \small
    \centering
    \caption{\textbf{Multiple sub-branch and photometric aug.} We test multiple sub-branch settings and additional photometric augmentation on sub-branch to further improve the performance. The result shows that a single sub-branch is the best for accuracy and computation. Additional color jittering on the sub-branch improves performance, but Gaussian blur is not effective.}
    \begin{tabular}{l|c|cc|cc}
    \toprule
    & MaskSub & + \begin{tabular}[c]{@{}c@{}}2-branch\\masking\end{tabular} & + \begin{tabular}[c]{@{}c@{}}3-branch\\masking\end{tabular} & + \begin{tabular}[c]{@{}c@{}}Color\\jitter\end{tabular} & + \begin{tabular}[c]{@{}c@{}}Gaussian\\blur\end{tabular} \\
    \midrule
    Accuracy & 84.10 & 84.03\redpscript{-0.07} & 83.76\redpscript{-0.34} & 84.18\greenpscript{+0.08} & 84.05\redpscript{-0.05} \\
    Costs & $\times$1.5 & $\times$2.0 & $\times$2.5 & $\times$1.5 & $\times$1.5\\
    \bottomrule
    \end{tabular}
    \label{atable:multi_photo}
\end{table}

\subsection{Reporting mean and standard deviation}
We provide mean and standard deviation for experiments using different random seeds. 
The values presented in this section result from three independent runs with different seeds.

\noindent Mean and standard deviation values for transfer learning in Table~\ref{table:transfer} are shown in Table~\ref{atable:transfer}. MaskSub demonstrates meaningful performance gains, which surpass the standard deviation of performance. Table~\ref{atable:transfer_short} presents short training (300 epochs) results. MaskSub shows substantial improvements when applied to pretraining and finetuning processes.  Table~\ref{atable:deit-iii_drop} shows 400 epochs training with DeiT-III~\cite{touvron2022deit3}, which is reported in Table~\ref{table:deit-iii_drop} of the paper. MaskSub variants improve the performance of ViT training. Compared to the variants, MaskSub demonstrates the best performance. Table~\ref{atable:analysis} shows the mean and standard deviation values for Table~\ref{table:analysis}. Table~\ref{atable:analysis} shows the superiority of our MaskSub in additional regularization training.

\begin{table*}[h]
\centering
\small
\setlength\tabcolsep{4pt}
\caption{\textbf{Mean and std for transfer learning to small scale datasets.} The table shows `mean $\pm$ std' values for transfer learning performance with/without MaskSub. We measure the performance when MaskSub is applied to pretraining and finetuning. }
\begin{tabular}{@{}ccccccccc@{}}
\toprule
\footnotesize Model & \begin{tabular}[c]{@{}c@{}}Pretraining\\+ MaskSub \end{tabular} & \begin{tabular}[c]{@{}c@{}}Finetuning\\+ MaskSub \end{tabular} & \footnotesize CIFAR10 & \footnotesize CIFAR100 & \footnotesize Flowers & \footnotesize Cars & \footnotesize iNat-18 & \footnotesize iNat-19\\ \midrule
\footnotesize \multirow{3}{*}[-0em]{ViT-S} & \footnotesize - & \footnotesize - &  
98.83 $\pm$ 0.05 & 89.96 $\pm$ 0.15 & 94.54 $\pm$ 1.71 & 80.86 $\pm$ 0.71 & 70.12 $\pm$ 0.13 & 76.69 $\pm$ 0.56 \\
& \footnotesize \textcolor{darkergreen}{\ding{52}}  & \footnotesize - &
98.88 $\pm$ 0.09 & 90.63 $\pm$ 0.09 & 95.19 $\pm$ 1.95 & 81.23 $\pm$ 0.73 & 70.82 $\pm$ 0.03 & 77.00 $\pm$ 0.21  \\ 
& \footnotesize \textcolor{darkergreen}{\ding{52}}  & \footnotesize \textcolor{darkergreen}{\ding{52}}  &
98.77 $\pm$ 0.05 & 89.87 $\pm$ 0.17 & 98.25 $\pm$ 0.51 &  92.17 $\pm$ 0.14 & 71.17 $\pm$ 0.21 & 77.12 $\pm$ 0.48   \\ \midrule
\footnotesize \multirow{3}{*}[-0em]{ViT-B} & \footnotesize - & \footnotesize - 
& 99.07 $\pm$ 0.05 & 91.69 $\pm$ 0.15 & 97.52 $\pm$ 0.51 & 90.05 $\pm$ 0.24 & 73.16 $\pm$ 0.05 & 78.49 $\pm$ 0.62\\
 & \footnotesize \textcolor{darkergreen}{\ding{52}}  & \footnotesize - &
 99.19 $\pm$ 0.03 & 91.89 $\pm$ 0.04 & 97.73 $\pm$ 0.30 & 90.18 $\pm$ 0.12 & 73.61 $\pm$ 0.08 & 78.77 $\pm$ 0.05\\ 
 & \footnotesize \textcolor{darkergreen}{\ding{52}}  & \footnotesize \textcolor{darkergreen}{\ding{52}}  &
 98.82 $\pm$ 0.03 & 89.55 $\pm$ 0.05 & 98.68 $\pm$ 0.16 & 92.77 $\pm$ 0.09 & 73.88 $\pm$ 0.12 & 79.07 $\pm$ 0.55 \\ \bottomrule
\end{tabular}
\label{atable:transfer}
\vspace{-1em}
\end{table*}

\begin{table*}[h]
\centering
\small
\setlength\tabcolsep{4pt}
\caption{\textbf{Transfer learning at short (300 epochs) training.} The table shows `mean $\pm$ std' values for transfer learning at 300 epochs. We measure the performance when MaskSub is applied to pretraining and finetuning. }
\begin{tabular}{@{}ccccccc@{}}
\toprule
Model & \begin{tabular}[c]{@{}c@{}}Pretraining\\+ MaskSub \end{tabular} & \begin{tabular}[c]{@{}c@{}}Finetuning\\+ MaskSub \end{tabular} & CIFAR10 & CIFAR100 & Flowers & Cars \\ \midrule
\multirow{3}{*}[-0em]{ViT-S} & - & - &  
98.41 $\pm$ 0.12 & 87.27 $\pm$ 0.19 & 66.66 $\pm$ 2.52 & 46.04 $\pm$ 3.72 \\
& \textcolor{darkergreen}{\ding{52}}  & - &
98.48 $\pm$ 0.06 & 87.54 $\pm$ 0.33 & 72.61 $\pm$ 1.20 & 45.46 $\pm$ 1.80 \\ 
& \textcolor{darkergreen}{\ding{52}}  & \textcolor{darkergreen}{\ding{52}}  &
\textbf{98.96} $\pm$ 0.06 & \textbf{90.81} $\pm$ 0.09 & \textbf{96.64} $\pm$ 0.23 & \textbf{87.43} $\pm$ 0.43 \\ \midrule
\multirow{3}{*}[-0em]{ViT-B} & - & - 
& 98.97 $\pm$ 0.10 & 90.33 $\pm$ 0.14 & 90.92 $\pm$ 1.60 & 78.52 $\pm$ 0.59 \\
& \textcolor{darkergreen}{\ding{52}}  & - &
 99.06 $\pm$ 0.02 & 90.82 $\pm$ 0.21 & 92.45 $\pm$ 0.90 & 80.34 $\pm$ 0.56 \\ 
& \textcolor{darkergreen}{\ding{52}}  & \textcolor{darkergreen}{\ding{52}}  &
 \textbf{99.15} $\pm$ 0.04 & \textbf{91.52} $\pm$ 0.16 & \textbf{98.44} $\pm$ 0.13 & \textbf{92.22} $\pm$ 0.03 \\ \bottomrule

\end{tabular}
\label{atable:transfer_short}
\vspace{-1em}
\end{table*}

\begin{table*}[h]
\centering
\small
\setlength\tabcolsep{5pt}
\caption{\textbf{Mean and std for analysis on drop regularization with/without MaskSub.} The table shows `mean $\pm$ std' values for experiments in Table~\ref{table:analysis} of the paper. Note that training loss scale $10^{-3}$ is omitted for simplicity. }
\begin{tabular}{@{}rccccccc@{}}
\toprule
         & & \multicolumn{3}{c}{Single model}  & \multicolumn{3}{c}{Sub-model training (MaskSub)} \\ \cmidrule(l){3-5} \cmidrule(l){6-8}
         & \begin{tabular}[c]{@{}c@{}}Drop\\ ratio\end{tabular} & Accuracy & \begin{tabular}[c]{@{}c@{}}Train loss\\(original)\end{tabular} & \begin{tabular}[c]{@{}c@{}}Train loss\\ (drop)\end{tabular} & Accuracy & \begin{tabular}[c]{@{}c@{}}Train loss\\ (original) \end{tabular} & \begin{tabular}[c]{@{}c@{}}Train loss\\ (drop)\end{tabular} \\ \cmidrule(l){1-2} \cmidrule(l){3-5} \cmidrule(l){6-8}
Original & - & 77.40 $\pm$ 0.20  &  6.42 $\pm$ 0.03 & -    &   -       & -   & -   \\ \cmidrule(r){1-2} \cmidrule(l){3-5} \cmidrule(l){6-8}
\multirow{3}{*}{Masking~\cite{he2022masked}} 
& 25\% & 76.33 $\pm$ 0.28 & 6.60 $\pm$ 0.05 & 6.96 $\pm$ 0.05  & 79.02 $\pm$ 0.12 & 5.89 $\pm$ 0.03  & 6.38 $\pm$ 0.04   \\ 
& 50\% & 73.78 $\pm$ 0.08  & 7.02 $\pm$ 0.04 & 7.77 $\pm$ 0.03   & 79.36 $\pm$ 0.01  & 5.81 $\pm$ 0.01  & 6.89 $\pm$ 0.01  \\  
& 75\% & 67.27 $\pm$ 0.25 & 8.08 $\pm$ 0.05 & 9.27 $\pm$ 0.04  & 79.16 $\pm$ 0.05 & 5.84 $\pm$ 0.01  & 8.15 $\pm$ 0.02   \\ \cmidrule(r){1-2} \cmidrule(l){3-5} \cmidrule(l){6-8}
\multirow{3}{*}{Drop-out~\cite{srivastava2014dropout}} 
& 0.1 & 76.09 $\pm$ 0.25 & 6.60 $\pm$ 0.07 & 6.87 $\pm$ 0.06  & 79.14 $\pm$ 0.15 & 5.88 $\pm$ 0.02  & 6.32 $\pm$ 0.02   \\ 
& 0.2 & 74.10 $\pm$ 0.22 & 6.95 $\pm$ 0.06 & 7.34 $\pm$ 0.06  & 79.10 $\pm$ 0.11 & 5.82 $\pm$ 0.04  & 6.57 $\pm$ 0.04   \\ 
& 0.3 & 71.62 $\pm$ 0.29 & 8.34 $\pm$ 0.03 & 7.79 $\pm$ 0.03  & 79.09 $\pm$ 0.15 & 5.84 $\pm$ 0.03  & 6.90 $\pm$ 0.03   \\ \cmidrule(r){1-2} \cmidrule(l){3-5} \cmidrule(l){6-8}
\multirow{3}{*}{Drop-path~\cite{huang2016stodepth}} 
& 0.1 & 77.40 $\pm$ 0.20 & 6.42 $\pm$ 0.03 & 6.42 $\pm$ 0.03  & 78.36 $\pm$ 0.03 & 6.11 $\pm$ 0.01  & 6.11 $\pm$ 0.01   \\ 
& 0.2 & 74.92 $\pm$ 0.12  &  6.74 $\pm$ 0.04 & 7.19 $\pm$ 0.03  & 78.72 $\pm$ 0.12  & 5.91 $\pm$ 0.01 & 6.48 $\pm$ 0.01 \\ 
& 0.3 & 71.57 $\pm$ 0.10 & 7.31 $\pm$ 0.02 & 8.04 $\pm$ 0.02  & 78.80 $\pm$ 0.15 & 5.87 $\pm$ 0.02 & 7.02 $\pm$ 0.01   \\ 
 \bottomrule
\end{tabular}
\label{atable:analysis}
\vspace{-1em}
\end{table*}

\begin{table*}[h]
\centering
\small
\setlength\tabcolsep{5pt}
\caption{\textbf{Mean and std for three variants of MaskSub.} We report `mean $\pm$ std' values for 400 epochs training with DeiT-III~\cite{touvron2022deit3}. Note that we use the performance of the original paper~\cite{touvron2022deit3} for baseline training.}
\begin{tabular}{@{}cccccc@{}}
\toprule
Architecture & Baseline & DropSub & PathSub & MaskSub  \\ \midrule
ViT-S/16   &   80.40 $\pm$ 0.33 & 80.57 $\pm$ 0.12 & 80.78 $\pm$ 0.05 & 81.08 $\pm$ 0.12 \\
ViT-B/16    &   83.46 $\pm$ 0.04 & 83.83 $\pm$ 0.11 & 83.80 $\pm$ 0.12 & 84.08 $\pm$ 0.02 \\ \midrule
\end{tabular}
\label{atable:deit-iii_drop}
\vspace{0em}
\end{table*}

\clearpage
\section{Implementation details}
\label{asubsec:impl_details}

Most experiments in the paper were performed on a machine with NVIDIA V100 8 GPUs. The exceptions were DeiT-III~\cite{touvron2022deit3} experiments for ViT-L and ViT-H in Table~\ref{table:deit-iii} 
and object detection in Table~\ref{table:detection}, conducted with NVIDIA A100 64 GPUs. Also, we use a single NVIDIA V100 for transfer learning, as shown in Table~\ref{table:transfer}. 
We strictly follow original training recipes for experiments. We denote details of the training recipes to clarify our implementation details and assist in reproducing our results. 

\noindent \textbf{Image Classification.} Table~\ref{atable:recipe} shows training recipes used for Table~\ref{table:deit-iii}, \ref{table:finetune}, \ref{table:resnet_swin}, \ref{table:comparison_deit}, \ref{table:deit-iii_drop} and \ref{table:comparison_finetune}. It demonstrates that our MaskSub is validated on various training recipes that cover diverse regularization and optimizer settings and achieves consistent improvement on all settings, which exhibits the general applicability of MaskSub. Note that MaskSub is applied to all recipes with the same masking ratio of 50\%. Thus, MaskSub does not require hyper-parameter tuning specialized for each recipe. Model-specific training recipes of DeiT-III~\cite{touvron2022deit3} are reported in Table~\ref{atable:deit_recipe}.
DeiT-III achieves strong performance with sophistically tuned training parameters mainly focused on input size and drop-path rate. It makes improving DeiT-III more challenging than other recipes, yet our MaskSub accomplishes this with a reasonable performance gap.

\noindent \textbf{Semantic segmentation.} We use the BEiT v2's segmentation recipe~\cite{peng2022beit2} that utilizes MMCV~\cite{mmcv} and MMSeg~\cite{mmseg2020} library. Following the default setting, we replace the ViT backbone with the DeiT-III backbone, which includes layer-scale~\cite{touvron2021going}. Then, we train the segmentation task for 160k iteration using DeiT-III and DeiT-III + MaskSub pretrained backbone.

\noindent \textbf{Object detection and instance segmentation} \noindent We use Detectron2~\cite{wu2019detectron2} on the COCO~\cite{lin2014mscoco} dataset. Among various recipes in the Detectron2 library, we use ViTDet~\cite{li2022vitdet}, which is a recent and strong recipe for a ViT-based detector. We train ViTDet Cascaded Mask-RCNN with DeiT-III and DeiT-III + MaskSub pretrained backbone and report performance after MSCOCO 100 epochs training.

\noindent \textbf{Transfer learning.} We use the AdamW training recipe on DeiT-III~\cite{touvron2022deit3} transfer learning. We use lr $10^{-5}$, weight-decay 0.05, batch size 768. Drop-path~\cite{huang2016stodepth} and random erasing~\cite{zhong2020randomerase} are not used. Data augmentation is set to be the same as DeiT-III, and we train ViT for 1000 epochs with a cosine learning rate decay. For CIFAR datasets, we resize $32\times32$ image to $224\times224$ to use the ImageNet pretrained backbones. In the case of the iNaturalist datasets~\cite{van2018inaturalist}, we use AdamW with lr $7.5 \times 10^{-5}$, weight-decay 0.05, batch size 768. Drop-path and random erasing ratios are set to 0.1, and ViT is trained for 360 epochs with cosine learning rate decay.

\noindent \textbf{Text classificaiton.} We use the HuggingFace open-source code for the GLUE benchmark~\cite{wang2018glue}. Random masking is implemented using the masking part of the HuggingFace BERT~\cite{devlin2018bert}, which replaces random tokens with a mask token. Following the BERT pretraining, we set a random masking ratio to 15\%, and only replacing tokens to the mask token is applied to randomly selected tokens.

\noindent \textbf{CLIP pretraining.} 
In pretraining, CLIP~\cite{clip}  encodes $N$ image-text pairs of a batch and builds an $ N\times N$ similarity matrix. Using contrastive learning loss, CLIP is trained to have high similarity for matched image-text pairs and low similarity for unmatched pairs.
Implementing MaskSub upon CLIP is straightforward since the contrastive learning loss also utilizes \textit{cross-entropy loss} over the similarity matrix. 
In the experiment, we only mask images, not texts. The difference between the ``main-model'' and ``sub-model'' is that the ``sub-model'' receives 50\% masked images, and the ``sub-model'' is trained to mimic the similarity matrix of the ``main-model'' for the same batch.
We use \textsc{open-clip}\footnote{\url{https://github.com/mlfoundations/open\_clip}}codebase for the CLIP pretraining. 
We follow the codebase's default training recipes. We train CLIP ViT-B/32 with a batch size of 1,024 using 8 NVIDIA V100 GPUs. 
Training datasets are CC3M \cite{sharma2018conceptual}, CC12M \cite{changpinyo2021cc12m}, and RedCaps \cite{desai2021redcaps}. The number of seen samples during training is 128M.

\begin{table}[h]
\centering
\small
\tabcolsep=0.17cm
\caption{\textbf{Details of various training recipes used for experiments.} Our MaskSub achieves consistent improvement in all training recipes, which covers diverse regularization and optimizer settings.}
\begin{tabular}{@{}lrrrrrr@{}}
\toprule
Training recipe & DeiT-III & RSB A2 & Swin & MAE & BEiT v2 & FT-CLIP \\ \midrule
Fine-tuning     & \textcolor{red}{\ding{55}}        & \textcolor{red}{\ding{55}}  & \textcolor{red}{\ding{55}}    & \textcolor{darkergreen}{\ding{52}}   & \textcolor{darkergreen}{\ding{52}}       & \textcolor{darkergreen}{\ding{52}}       \\ \midrule
Epoch   &   400 / 800  &  300   &  300  & 100, 50 & 100, 50 & 50, 30  \\
Batch size & 2048 & 2048 & 1024 & 1024 & 1024 & 2048 \\
Optimizer & LAMB  & LAMB & AdamW & AdamW & AdamW & AdamW \\
LR & $3\times 10^{-3}$  & $5\times 10^{-3}$  & $1\times 10^{-3}$ & $(2, 4)\times 10^{-3}$  & $(5, 2)\times 10^{-4}$ & $(6, 4)\times 10^{-4}$\\
LR decay & cosine & cosine & cosine & cosine & cosine & cosine \\
Layer LR decay & - & -  & - & 0.65, 0.75  & 0.6, 0.8 & 0.6, 0.65\\ 
Weight decay & 0.03 / 0.05 & 0.01  & 0.05 & 0.05  & 0.05 & 0.05 \\
Warmup epochs & 5 & 5  & 20 & 5  & 20, 5 & 10, 5\\ 
Loss & BCE & BCE & CE & CE & CE & CE\\ \midrule
Label smoothing & - & - & 0.1  & 0.1 & 0.1 & 0.1\\
Dropout & - & - & - & - & - & - \\
Drop-path & Table~\ref{atable:deit_recipe} & 0.05 & 0.1 & 0.1, 0.2, 0.3 & 0.2 & -\\
Repeated aug & \textcolor{darkergreen}{\ding{52}} & \textcolor{darkergreen}{\ding{52}} & - & - & - & - \\
Gradient clip & 1.0 & - & 5.0 & - & - & - \\
RandAugment & Three Aug. & 7 / 0.5 & 9 / 0.5  & 9 / 0.5 & 9 / 0.5 & 9 / 0.5 \\
Mixup alpha & 0.8 & 0.1 & 0.8 & 0.8 & 0.8 & - \\
CutMix alpha & 1.0 & 1.0 & 1.0 & 1.0 & 1.0 & - \\
Random erasing & - & - & 0.25 & 0.25 & 0.25 & 0.25 \\
Color jitter & 0.3 & - & 0.4 & - & 0.4 & 0.4 \\ 
EMA & - & - & - & - & - & 0.9998\\ 
Train image size & Table~\ref{atable:deit_recipe} & $224\times224$ & $224\times224$  & $224\times224$ & $224\times224$  &  $224\times224$ \\ \midrule
Test image size & $224\times224$ & $224\times224$  & $224\times224$ & $224\times224$ & $224\times224$  & $224\times224$ \\ 
Test crop ratio & 1.0 & 0.95  & 0.875  & 0.875 & 0.875 & 1.0 \\ \bottomrule
\end{tabular}
\label{atable:recipe}
\end{table}

\begin{table}[h]
\centering
\small
\setlength\tabcolsep{5pt}
\caption{\textbf{Model specific recipes of DeiT-III~\cite{touvron2022deit3}.} The table shows the model size and training length-specific training arguments used for the DeiT-III recipe. In addition to Table~\ref{atable:recipe}, DeiT-III utilizes drop-path and image size to adjust the recipe.}
\begin{tabular}{@{}crcccccccc@{}}
\toprule
             & & \multicolumn{4}{c}{400 epochs}         & \multicolumn{4}{c}{800 epochs}         \\ \cmidrule(l){3-6}  \cmidrule(l){7-10}
             & & ViT-S  & ViT-B     & ViT-L   & ViT-H   & ViT-S  & ViT-B     & ViT-L   & ViT-H   \\ 
             \cmidrule(l){1-2} \cmidrule(l){3-3} \cmidrule(l){4-4} \cmidrule(l){5-5} \cmidrule(l){6-6} \cmidrule(l){7-7} \cmidrule(l){8-8} \cmidrule(l){9-9} \cmidrule(l){10-10} 
\multirow{4}{*}[-0.4em]{Pretraining} 
& Image size & 224    & 192       & 192     & 160     & 224    & 192       & 192     & 160     \\ 
& Drop-path     & 0.0    & 0.1       & 0.4     & 0.5     & 0.05   & 0.2       & 0.45    & 0.6     \\ \cmidrule(l){3-3} \cmidrule(l){4-6} \cmidrule(l){7-7} \cmidrule(l){8-10}
& LR & 0.004  & \multicolumn{3}{c}{0.003}     & 0.004  & \multicolumn{3}{c}{0.003}     \\ \cmidrule(l){3-6} \cmidrule(l){7-10}
& Weight decay  & \multicolumn{4}{c}{0.03}               & \multicolumn{4}{c}{0.05}               \\ \cmidrule(l){1-2} \cmidrule(l){3-3} \cmidrule(l){4-4} \cmidrule(l){5-5} \cmidrule(l){6-6} \cmidrule(l){7-7} \cmidrule(l){8-8} \cmidrule(l){9-9} \cmidrule(l){10-10} 
\multirow{4}{*}[-0.7em]{\begin{tabular}[c]{@{}c@{}}Resolution\\ Finetuning\end{tabular}} & Drop-path     & -      & 0.2       & 0.45    & 0.55    & -      & 0.2       & 0.45    & 0.55     \\ \cmidrule(l){3-3} \cmidrule(l){4-6} \cmidrule(l){7-7} \cmidrule(l){8-10}
& Epochs        & -      & \multicolumn{3}{c}{20}        & -      & \multicolumn{3}{c}{20}        \\
& Image size    & -      & \multicolumn{3}{c}{224 x 224} & -      & \multicolumn{3}{c}{224 x 224} \\
& Optimizer     & -      & \multicolumn{3}{c}{AdamW}     & -      & \multicolumn{3}{c}{AdamW}     \\
& LR & -      & \multicolumn{3}{c}{1e-5}      & -      & \multicolumn{3}{c}{1e-5}      \\   \bottomrule
\end{tabular}
\label{atable:deit_recipe}
\end{table}

\end{document}